\documentclass{Interspeech}
\interspeechcameraready

\usepackage[utf8]{inputenc}
\usepackage[T1]{fontenc}
\usepackage{xcolor}
\usepackage{microtype}
\usepackage{pgfplots}
\pgfplotsset{compat=1.17}
\usepgfplotslibrary{groupplots}

\input{results/numbers}
\expandafter\def\csname res@cal-3way-k0-linear-ctx-strict\endcsname{62.6}
\expandafter\def\csname res@cal-3way-k0-argmax\endcsname{58.6}
\expandafter\def\csname res@cal-3way-k0-argmax-vs-probe\endcsname{-3.9}
\expandafter\def\csname res@cal-3way-k0-argmax-vs-probe-cilo\endcsname{-4.3}
\expandafter\def\csname res@cal-3way-k0-argmax-vs-probe-cihi\endcsname{-3.6}
\expandafter\def\csname res@cal-3way-k0-logistic\endcsname{59.6}
\expandafter\def\csname res@cal-3way-k0-logistic-vs-probe\endcsname{-3.0}
\expandafter\def\csname res@cal-3way-k0-logistic-vs-probe-cilo\endcsname{-3.3}
\expandafter\def\csname res@cal-3way-k0-logistic-vs-probe-cihi\endcsname{-2.6}
\expandafter\def\csname res@cal-3way-k0-cutoffs\endcsname{59.6}
\expandafter\def\csname res@cal-3way-k0-cutoffs-vs-probe\endcsname{-3.0}
\expandafter\def\csname res@cal-3way-k0-cutoffs-vs-probe-cilo\endcsname{-3.3}
\expandafter\def\csname res@cal-3way-k0-cutoffs-vs-probe-cihi\endcsname{-2.6}
\expandafter\def\csname res@cal-3way-k0-linear-emb\endcsname{60.7}
\expandafter\def\csname res@cal-3way-k0-linear-emb-vs-probe\endcsname{-1.9}
\expandafter\def\csname res@cal-3way-k0-linear-emb-vs-probe-cilo\endcsname{-2.2}
\expandafter\def\csname res@cal-3way-k0-linear-emb-vs-probe-cihi\endcsname{-1.6}
\expandafter\def\csname res@cal-3way-k0-linear-rand\endcsname{58.4}
\expandafter\def\csname res@cal-3way-k0-linear-rand-vs-probe\endcsname{-4.1}
\expandafter\def\csname res@cal-3way-k0-linear-rand-vs-probe-cilo\endcsname{-4.5}
\expandafter\def\csname res@cal-3way-k0-linear-rand-vs-probe-cihi\endcsname{-3.8}
\expandafter\def\csname res@cal-3way-k0-lookup\endcsname{52.9}
\expandafter\def\csname res@cal-3way-k0-lookup-vs-probe\endcsname{-9.6}
\expandafter\def\csname res@cal-3way-k0-lookup-vs-probe-cilo\endcsname{-10.0}
\expandafter\def\csname res@cal-3way-k0-lookup-vs-probe-cihi\endcsname{-9.3}
\expandafter\def\csname res@cal-3way-k500-linear-ctx-strict\endcsname{62.6}
\expandafter\def\csname res@cal-3way-k500-argmax\endcsname{62.0}
\expandafter\def\csname res@cal-3way-k500-argmax-vs-probe\endcsname{-0.6}
\expandafter\def\csname res@cal-3way-k500-argmax-vs-probe-cilo\endcsname{-0.9}
\expandafter\def\csname res@cal-3way-k500-argmax-vs-probe-cihi\endcsname{-0.2}
\expandafter\def\csname res@cal-3way-k500-logistic\endcsname{62.3}
\expandafter\def\csname res@cal-3way-k500-logistic-vs-probe\endcsname{-0.2}
\expandafter\def\csname res@cal-3way-k500-logistic-vs-probe-cilo\endcsname{-0.6}
\expandafter\def\csname res@cal-3way-k500-logistic-vs-probe-cihi\endcsname{0.1}
\expandafter\def\csname res@cal-3way-k500-cutoffs\endcsname{62.0}
\expandafter\def\csname res@cal-3way-k500-cutoffs-vs-probe\endcsname{-0.5}
\expandafter\def\csname res@cal-3way-k500-cutoffs-vs-probe-cilo\endcsname{-0.8}
\expandafter\def\csname res@cal-3way-k500-cutoffs-vs-probe-cihi\endcsname{-0.2}
\expandafter\def\csname res@cal-3way-k500-linear-emb\endcsname{60.7}
\expandafter\def\csname res@cal-3way-k500-linear-emb-vs-probe\endcsname{-1.9}
\expandafter\def\csname res@cal-3way-k500-linear-emb-vs-probe-cilo\endcsname{-2.2}
\expandafter\def\csname res@cal-3way-k500-linear-emb-vs-probe-cihi\endcsname{-1.6}
\expandafter\def\csname res@cal-3way-k500-linear-rand\endcsname{58.4}
\expandafter\def\csname res@cal-3way-k500-linear-rand-vs-probe\endcsname{-4.1}
\expandafter\def\csname res@cal-3way-k500-linear-rand-vs-probe-cilo\endcsname{-4.5}
\expandafter\def\csname res@cal-3way-k500-linear-rand-vs-probe-cihi\endcsname{-3.8}
\expandafter\def\csname res@cal-3way-k500-lookup\endcsname{52.9}
\expandafter\def\csname res@cal-3way-k500-lookup-vs-probe\endcsname{-9.6}
\expandafter\def\csname res@cal-3way-k500-lookup-vs-probe-cilo\endcsname{-10.0}
\expandafter\def\csname res@cal-3way-k500-lookup-vs-probe-cihi\endcsname{-9.3}
\expandafter\def\csname res@cal-2way-k0-linear-ctx-strict\endcsname{81.1}
\expandafter\def\csname res@cal-2way-k0-argmax\endcsname{79.0}
\expandafter\def\csname res@cal-2way-k0-argmax-vs-probe\endcsname{-2.2}
\expandafter\def\csname res@cal-2way-k0-argmax-vs-probe-cilo\endcsname{-2.4}
\expandafter\def\csname res@cal-2way-k0-argmax-vs-probe-cihi\endcsname{-1.9}
\expandafter\def\csname res@cal-2way-k0-logistic\endcsname{79.4}
\expandafter\def\csname res@cal-2way-k0-logistic-vs-probe\endcsname{-1.8}
\expandafter\def\csname res@cal-2way-k0-logistic-vs-probe-cilo\endcsname{-2.0}
\expandafter\def\csname res@cal-2way-k0-logistic-vs-probe-cihi\endcsname{-1.6}
\expandafter\def\csname res@cal-2way-k0-cutoffs\endcsname{79.4}
\expandafter\def\csname res@cal-2way-k0-cutoffs-vs-probe\endcsname{-1.7}
\expandafter\def\csname res@cal-2way-k0-cutoffs-vs-probe-cilo\endcsname{-1.9}
\expandafter\def\csname res@cal-2way-k0-cutoffs-vs-probe-cihi\endcsname{-1.5}
\expandafter\def\csname res@cal-2way-k0-linear-emb\endcsname{80.7}
\expandafter\def\csname res@cal-2way-k0-linear-emb-vs-probe\endcsname{-0.5}
\expandafter\def\csname res@cal-2way-k0-linear-emb-vs-probe-cilo\endcsname{-0.7}
\expandafter\def\csname res@cal-2way-k0-linear-emb-vs-probe-cihi\endcsname{-0.3}
\expandafter\def\csname res@cal-2way-k0-linear-rand\endcsname{79.2}
\expandafter\def\csname res@cal-2way-k0-linear-rand-vs-probe\endcsname{-1.9}
\expandafter\def\csname res@cal-2way-k0-linear-rand-vs-probe-cilo\endcsname{-2.1}
\expandafter\def\csname res@cal-2way-k0-linear-rand-vs-probe-cihi\endcsname{-1.7}
\expandafter\def\csname res@cal-2way-k0-lookup\endcsname{79.4}
\expandafter\def\csname res@cal-2way-k0-lookup-vs-probe\endcsname{-1.8}
\expandafter\def\csname res@cal-2way-k0-lookup-vs-probe-cilo\endcsname{-2.0}
\expandafter\def\csname res@cal-2way-k0-lookup-vs-probe-cihi\endcsname{-1.6}
\expandafter\def\csname res@cal-2way-k500-linear-ctx-strict\endcsname{81.1}
\expandafter\def\csname res@cal-2way-k500-argmax\endcsname{79.5}
\expandafter\def\csname res@cal-2way-k500-argmax-vs-probe\endcsname{-1.6}
\expandafter\def\csname res@cal-2way-k500-argmax-vs-probe-cilo\endcsname{-1.9}
\expandafter\def\csname res@cal-2way-k500-argmax-vs-probe-cihi\endcsname{-1.4}
\expandafter\def\csname res@cal-2way-k500-logistic\endcsname{80.0}
\expandafter\def\csname res@cal-2way-k500-logistic-vs-probe\endcsname{-1.1}
\expandafter\def\csname res@cal-2way-k500-logistic-vs-probe-cilo\endcsname{-1.3}
\expandafter\def\csname res@cal-2way-k500-logistic-vs-probe-cihi\endcsname{-0.9}
\expandafter\def\csname res@cal-2way-k500-cutoffs\endcsname{80.0}
\expandafter\def\csname res@cal-2way-k500-cutoffs-vs-probe\endcsname{-1.1}
\expandafter\def\csname res@cal-2way-k500-cutoffs-vs-probe-cilo\endcsname{-1.3}
\expandafter\def\csname res@cal-2way-k500-cutoffs-vs-probe-cihi\endcsname{-0.9}
\expandafter\def\csname res@cal-2way-k500-linear-emb\endcsname{80.7}
\expandafter\def\csname res@cal-2way-k500-linear-emb-vs-probe\endcsname{-0.5}
\expandafter\def\csname res@cal-2way-k500-linear-emb-vs-probe-cilo\endcsname{-0.7}
\expandafter\def\csname res@cal-2way-k500-linear-emb-vs-probe-cihi\endcsname{-0.3}
\expandafter\def\csname res@cal-2way-k500-linear-rand\endcsname{79.2}
\expandafter\def\csname res@cal-2way-k500-linear-rand-vs-probe\endcsname{-1.9}
\expandafter\def\csname res@cal-2way-k500-linear-rand-vs-probe-cilo\endcsname{-2.1}
\expandafter\def\csname res@cal-2way-k500-linear-rand-vs-probe-cihi\endcsname{-1.7}
\expandafter\def\csname res@cal-2way-k500-lookup\endcsname{79.4}
\expandafter\def\csname res@cal-2way-k500-lookup-vs-probe\endcsname{-1.8}
\expandafter\def\csname res@cal-2way-k500-lookup-vs-probe-cilo\endcsname{-2.0}
\expandafter\def\csname res@cal-2way-k500-lookup-vs-probe-cihi\endcsname{-1.6}

\expandafter\def\csname res@v2-e2a-agree-probe\endcsname{81.7}
\expandafter\def\csname res@v2-e2a-agree-prompt\endcsname{72.7}
\expandafter\def\csname res@v2-e2a-agree-text\endcsname{82.8}
\expandafter\def\csname res@v2-e2a-agree-words\endcsname{1,224}
\expandafter\def\csname res@v2-e2a-boundary-gap-share\endcsname{23}
\expandafter\def\csname res@v2-e2a-boundary-word-share\endcsname{14}
\expandafter\def\csname res@v2-e2a-differ-words\endcsname{682}
\expandafter\def\csname res@v2-e2a-gap-final\endcsname{7.9}
\expandafter\def\csname res@v2-e2a-gap-other\endcsname{4.2}
\expandafter\def\csname res@v2-e2b-ac0-2\endcsname{79.1}
\expandafter\def\csname res@v2-e2b-ac0-2-cihi\endcsname{-0.3}
\expandafter\def\csname res@v2-e2b-ac0-2-cilo\endcsname{-1.1}
\expandafter\def\csname res@v2-e2b-ac0-2-diff\endcsname{-0.7}
\expandafter\def\csname res@v2-e2b-ac0-2-ref\endcsname{79.8}
\expandafter\def\csname res@v2-e2b-ac0-3\endcsname{60.2}
\expandafter\def\csname res@v2-e2b-ac0-3-cihi\endcsname{0.3}
\expandafter\def\csname res@v2-e2b-ac0-3-cilo\endcsname{-0.8}
\expandafter\def\csname res@v2-e2b-ac0-3-diff\endcsname{-0.3}
\expandafter\def\csname res@v2-e2b-ac0-3-ref\endcsname{60.5}
\expandafter\def\csname res@v2-e2b-ac500-2\endcsname{79.8}
\expandafter\def\csname res@v2-e2b-ac500-2-cihi\endcsname{0.5}
\expandafter\def\csname res@v2-e2b-ac500-2-cilo\endcsname{-0.4}
\expandafter\def\csname res@v2-e2b-ac500-2-diff\endcsname{0.0}
\expandafter\def\csname res@v2-e2b-ac500-2-ref\endcsname{79.8}
\expandafter\def\csname res@v2-e2b-ac500-3\endcsname{63.0}
\expandafter\def\csname res@v2-e2b-ac500-3-cihi\endcsname{0.0}
\expandafter\def\csname res@v2-e2b-ac500-3-cilo\endcsname{-1.1}
\expandafter\def\csname res@v2-e2b-ac500-3-diff\endcsname{-0.5}
\expandafter\def\csname res@v2-e2b-ac500-3-ref\endcsname{63.5}
\expandafter\def\csname res@v2-e2b-pw-2\endcsname{79.2}
\expandafter\def\csname res@v2-e2b-pw-2-cihi\endcsname{-0.1}
\expandafter\def\csname res@v2-e2b-pw-2-cilo\endcsname{-1.1}
\expandafter\def\csname res@v2-e2b-pw-2-diff\endcsname{-0.6}
\expandafter\def\csname res@v2-e2b-pw-2-ref\endcsname{79.8}
\expandafter\def\csname res@v2-e2b-pw-3\endcsname{59.9}
\expandafter\def\csname res@v2-e2b-pw-3-cihi\endcsname{-0.3}
\expandafter\def\csname res@v2-e2b-pw-3-cilo\endcsname{-1.8}
\expandafter\def\csname res@v2-e2b-pw-3-diff\endcsname{-1.0}
\expandafter\def\csname res@v2-e2b-pw-3-ref\endcsname{60.9}
\expandafter\def\csname res@v2-e3-probe-prompt\endcsname{4.7}
\expandafter\def\csname res@v2-e3-probe-prompt-cihi\endcsname{5.0}
\expandafter\def\csname res@v2-e3-probe-prompt-cilo\endcsname{4.4}
\expandafter\def\csname res@v2-e3-reader-speech\endcsname{14.5}
\expandafter\def\csname res@v2-e3-reader-speech-cihi\endcsname{17.0}
\expandafter\def\csname res@v2-e3-reader-speech-cilo\endcsname{11.9}
\expandafter\def\csname res@v2-e3-speech-text\endcsname{16.5}
\expandafter\def\csname res@v2-e3-speech-text-cihi\endcsname{16.9}
\expandafter\def\csname res@v2-e3-speech-text-cilo\endcsname{16.2}
\expandafter\def\csname res@v2-e4-distance-full\endcsname{12.3}
\expandafter\def\csname res@v2-e4-distance-observed\endcsname{11.8}
\expandafter\def\csname res@v2-e4-long-high\endcsname{79.5}
\expandafter\def\csname res@v2-e4-long-text\endcsname{68.8}
\expandafter\def\csname res@v2-e4-noise-full-err\endcsname{10.2}
\expandafter\def\csname res@v2-e4-noise-full-high\endcsname{83.5}
\expandafter\def\csname res@v2-e4-noise-full-text\endcsname{71.1}
\expandafter\def\csname res@v2-e4-short-high\endcsname{80.8}
\expandafter\def\csname res@v2-e4-short-text\endcsname{64.8}
\expandafter\def\csname res@v2-e4-spk-high\endcsname{80.4}
\expandafter\def\csname res@v2-e4-spk-low\endcsname{64.6}
\expandafter\def\csname res@v2-sat-k0-2-n100\endcsname{79.5}
\expandafter\def\csname res@v2-sat-k0-2-n250\endcsname{79.4}
\expandafter\def\csname res@v2-sat-k0-2-n50\endcsname{79.3}
\expandafter\def\csname res@v2-sat-k0-2-n500\endcsname{79.4}
\expandafter\def\csname res@v2-sat-k0-2-own\endcsname{79.0}
\expandafter\def\csname res@v2-sat-k0-3-n100\endcsname{59.6}
\expandafter\def\csname res@v2-sat-k0-3-n250\endcsname{59.5}
\expandafter\def\csname res@v2-sat-k0-3-n50\endcsname{59.4}
\expandafter\def\csname res@v2-sat-k0-3-n500\endcsname{59.6}
\expandafter\def\csname res@v2-sat-k0-3-own\endcsname{58.6}
\expandafter\def\csname res@v2-sat-k500-2-n100\endcsname{80.0}
\expandafter\def\csname res@v2-sat-k500-2-n250\endcsname{80.0}
\expandafter\def\csname res@v2-sat-k500-2-n50\endcsname{80.0}
\expandafter\def\csname res@v2-sat-k500-2-n500\endcsname{80.0}
\expandafter\def\csname res@v2-sat-k500-2-own\endcsname{79.5}
\expandafter\def\csname res@v2-sat-k500-3-n100\endcsname{62.2}
\expandafter\def\csname res@v2-sat-k500-3-n250\endcsname{62.3}
\expandafter\def\csname res@v2-sat-k500-3-n50\endcsname{62.0}
\expandafter\def\csname res@v2-sat-k500-3-n500\endcsname{62.3}
\expandafter\def\csname res@v2-sat-k500-3-own\endcsname{62.0}
\expandafter\def\csname res@v2-wt-emb\endcsname{0.52}
\expandafter\def\csname res@v2-wt-emb-nb\endcsname{0.52}
\expandafter\def\csname res@v2-wt-gain\endcsname{0.10}
\expandafter\def\csname res@v2-wt-gain-cihi\endcsname{0.12}
\expandafter\def\csname res@v2-wt-gain-cilo\endcsname{0.09}
\expandafter\def\csname res@v2-wt-gain-nb\endcsname{0.10}
\expandafter\def\csname res@v2-wt-gain-nb-cihi\endcsname{0.11}
\expandafter\def\csname res@v2-wt-gain-nb-cilo\endcsname{0.08}
\expandafter\def\csname res@v2-wt-p0\endcsname{0.53}
\expandafter\def\csname res@v2-wt-p0-nb\endcsname{0.53}
\expandafter\def\csname res@v2-wt-p500\endcsname{0.63}
\expandafter\def\csname res@v2-wt-p500-minus-pos\endcsname{0.09}
\expandafter\def\csname res@v2-wt-p500-minus-pos-cihi\endcsname{0.11}
\expandafter\def\csname res@v2-wt-p500-minus-pos-cilo\endcsname{0.07}
\expandafter\def\csname res@v2-wt-p500-minus-pos-nb\endcsname{0.04}
\expandafter\def\csname res@v2-wt-p500-minus-pos-nb-cihi\endcsname{0.06}
\expandafter\def\csname res@v2-wt-p500-minus-pos-nb-cilo\endcsname{0.03}
\expandafter\def\csname res@v2-wt-p500-nb\endcsname{0.62}
\expandafter\def\csname res@v2-wt-p500rec\endcsname{0.63}
\expandafter\def\csname res@v2-wt-p500rec-nb\endcsname{0.62}
\expandafter\def\csname res@v2-wt-pos\endcsname{0.54}
\expandafter\def\csname res@v2-wt-pos-nb\endcsname{0.58}
\expandafter\def\csname res@v2-wt-probe\endcsname{0.69}
\expandafter\def\csname res@v2-wt-probe-minus-p500\endcsname{0.06}
\expandafter\def\csname res@v2-wt-probe-minus-p500-cihi\endcsname{0.08}
\expandafter\def\csname res@v2-wt-probe-minus-p500-cilo\endcsname{0.05}
\expandafter\def\csname res@v2-wt-probe-minus-p500-nb\endcsname{0.07}
\expandafter\def\csname res@v2-wt-probe-minus-p500-nb-cihi\endcsname{0.09}
\expandafter\def\csname res@v2-wt-probe-minus-p500-nb-cilo\endcsname{0.06}
\expandafter\def\csname res@v2-wt-probe-nb\endcsname{0.69}
\expandafter\def\csname res@v2-wt-share\endcsname{75}
\expandafter\def\csname res@v2-wt-share-nb\endcsname{76}
\expandafter\def\csname res@v2-wt-share-of-probe\endcsname{63}
\expandafter\def\csname res@v2-wt-share-of-probe-nb\endcsname{58}
\expandafter\def\csname res@v2-wt-types\endcsname{948}
\expandafter\def\csname res@v2-wt-types-nb\endcsname{711}

\title{Language models represent more about emphasis than they can tell you}
\author[affiliation={1}]{Aarne}{Talman}
\author[affiliation={2,3}]{Stergios}{Chatzikyriakidis}
\affiliation{}{University of Helsinki}{Finland}
\affiliation{}{University of Turku}{Finland}
\affiliation{Institute for Language and Speech Processing}{Athena Research Center}{Greece}
\email{aarne.talman@helsinki.fi, s.chatzikyriakidis@athenarc.gr}
\keywords{prosodic prominence, large language models, probing, prompting, speech representations}

\begin{document}
\setlength{\textfloatsep}{6pt plus 2pt minus 2pt}
\setlength{\floatsep}{6pt plus 1pt minus 1pt}
\setlength{\intextsep}{6pt plus 1pt minus 1pt}
\setlength{\dbltextfloatsep}{6pt plus 2pt minus 2pt}
\setlength{\dblfloatsep}{6pt plus 1pt minus 1pt}
\setlength{\abovecaptionskip}{4pt plus 1pt minus 1pt}
\setlength{\belowcaptionskip}{2pt plus 1pt minus 1pt}
\maketitle

\begin{abstract}
Prosodic prominence can largely be predicted from text, but part of it depends on how the speaker reads the sentence.
We examine how much information about emphasis language models recover from text, whether they can report it when asked, and what the speech signal adds.
Using English speech data with automatically derived word-level emphasis labels, we compare frozen and fine-tuned text models, prompted large language models (LLMs) and speech models on the same words.
The text models end within about one point of each other, at best \resval{text-best-3}\% on the 3-way task, and make many of the same errors.
With the prompts we tested, the LLMs do no better than a word-majority baseline, even with 500 examples.
While LLM responses contain some contextual information, they contain less than linear probes find in the model representations, and the gap does not close when the answers are recalibrated or the question is asked differently.
The best speech model reaches \resval{p-whisper-bilstm-3}\%.
We argue that text determines only part of the emphasis, and that the rest has to be heard.
LLMs encode what the text determines, but prompting reveals only part of it: LLMs represent more about where the emphasis falls than they tell us.
\end{abstract}

\section{Introduction}

When a sentence is read aloud, some of its words stand out more than others.
Following Talman et al.~\cite{talman2019}, we regard this prosodic prominence not as contrastive or emphatic focus in the narrow sense, but as the degree of emphasis the speaker assigns to the words.
There is no formal definition of prominence that is universally accepted across languages, and the debate over whether prominence at the word level and the phrase level is a single phenomenon or two is still ongoing \cite{ladd2023}, so our definition is operational: prominence is what the labels measure.
Much of this is predictable from the text.
Content words are usually more emphasised than function words are, new information more than given information, and the last content word of a phrase tends to carry the main accent \cite{hirschberg1993,wagner2010}.
However, emphasis can also vary depending on what the speaker intends to convey.
Hence Bolinger's remark that accent is predictable only if you are a mind-reader \cite{bolinger1972}.
Speakers agree on accent placement only some of the time \cite{yuan2005}.

Large language models (LLMs) are now used to choose which words to emphasise in speech synthesis \cite{li2025}, continuing a line of work that uses text representations to predict prominence in synthesis front-ends \cite{rendel2016,stephenson2022}.
It is reasonable to expect that models trained on large amounts of text have learned much about information structure, and therefore about prominence.
Do their representations encode where the emphasis falls?
And if they do, can they tell us?

In this paper we answer these questions with a comparison of text and speech models on the prominence annotations of the LibriTTS audiobooks released by Talman et al.~\cite{talman2019}.
Since the labels were computed automatically from the recordings, text models, prompted LLMs and speech models can be evaluated on the same labels and the same words.
The main contributions of this study are as follows:
\begin{itemize}
\item \textbf{LLMs represent more about prominence than they can report.} A linear probe on the hidden states of Qwen3.8 27B demonstrated an accuracy of \resval{p-qwen-lin-3}\% in a three-way task, which is higher than the \resval{emb-qwen-3}\% achieved by the same probe on its embedding layer, which encodes each word without context. Therefore, contextual representations contain information about prominence beyond the identity of the word. For the same labels, when using the prompts tested in this study, the model demonstrated lower accuracy. It showed an accuracy of \resval{icl-test-k0-3}\% when there were no examples in the prompt and \resval{icl-test-3}\% when there were 500 examples. This is no better than a word-majority baseline that gives each word the label it most often has in the training data. Recalibrating the model's own answer does not change this, and within words that are prominent in some contexts and not in others, its answer separates the two less well than the probe does.
\item \textbf{One ceiling for all text models.} Text models of different sizes and designs, probed or fine-tuned, converge on the same accuracy and largely on the same errors, and a classifier over random word vectors comes close to it, which suggests that, for the models and data we tested, the limit lies in the text rather than in the models.
\item \textbf{Benefits provided by speech signals.} The best speech model, which hears how a sentence was actually read, is \resval{gap-3} points more accurate than the best text model for the same words. This improvement in accuracy is most pronounced in content words and the last word of the utterance, and when two speakers read the same text, the speech model follows the speaker it hears.
\item \textbf{A ceiling for speech models.} The labels were computed automatically from the recordings. When the same procedure is run again on public word alignments, the original labels are reproduced for \resval{toolkit-3}\% of the words (3-way). This shows how far a speech model could go on these labels. The best speech model is still \resval{toolkit-gap-3} points below this reproduction.
\end{itemize}
\ifinterspeechfinal
The code, the results of every run and the commands that reproduce them are publicly available.\footnote{\url{https://github.com/aarnetalman/prosody-llms}}
\fi

\section{Related work}

Prominence and pitch accents have been predicted from text with decision trees and classifiers over part of speech, word class and information status \cite{hirschberg1993}, with conditional random fields \cite{gregory2004}, from lexical statistics in conversational speech \cite{nenkova2007}, and with neural networks over word embeddings \cite{stehwien2018,rendel2016}.
Nielsen et al.~\cite{nielsen2020} show that an acoustic pitch-accent detector gains from the context of the whole utterance, and that accenting every content word is already a strong baseline (82.2\% accuracy on the Boston University Radio News Corpus), a point that our word-majority baseline echoes.
Talman et al.~\cite{talman2019} annotated LibriTTS for prominence and showed that fine-tuned BERT \cite{devlin2019} outperformed recurrent and feature-based models on the task.
Probing the layers of frozen BERT on these labels, Kakouros and O'Mahony \cite{kakouros2023bert} found information about prominence mostly in the middle layers, and the surprisal of words under LLMs correlates with prominence, although it helps speech synthesis only a little \cite{kakouros2023surprisal}.
In speech, prominence has been detected both with supervised classifiers and with unsupervised methods, among them the wavelet-based method of Suni et al.~\cite{suni2017}, which was used to produce these labels, and 3PRO \cite{kakouros2016}.
Self-supervised speech models \cite{hsu2021,chen2022}, speech recognisers \cite{radford2023} and speech-text LLMs such as Qwen2-Audio\footnote{Y.~Chu et al., Qwen2-Audio technical report, arXiv:2407.10759, 2024.} now provide representations of the signal that can be probed in the same way as text representations.
Layer-wise probing has found suprasegmental information, including English phrasal pitch accents, mostly in the middle layers of such models \cite{delafuente2024}, and prosodic structure, among it prominence, can be predicted from self-supervised representations of pitch, energy and timing alone \cite{wallbridge2025}.
Sentence stress can be reliably detected by a Whisper-based model \cite{yosha2025whistress}.
However, speech language models, Qwen2-Audio among them, perform close to chance on the same task \cite{yosha2026stresstest}.
Audio language models have also been shown to encode prosodic and paralinguistic cues that are lost before the output or not used in their answers.\footnote{L.~Peng and B.~Nuchged, \emph{Represented but ignored: a causal account of prosodic underuse in audio-language models}, arXiv:2608.19211, 2026; B.~Koduru et al., \emph{Heard but not heeded: paralinguistic information encoding and loss in audio-language models}, arXiv:2609.00727, 2026.}
Our speech results are consistent with these findings and extend the comparison to text models and prompted LLMs on the same labels.

Probing classifiers measure what information frozen representations provide \cite{belinkov2022}.
However, powerful probes can learn the task themselves \cite{hewitt2019}, so we compare our classifiers with a control representation of random word vectors and base our claims about what representations contain on linear probes.
In-context learning \cite{brown2020} with many examples has been shown to approach fine-tuning on some tasks \cite{agarwal2024}.
Hu and Levy \cite{hu2023} show that LLMs' answers to metalinguistic prompts can underestimate what is encoded in their probability distributions, and Hu and Frank \cite{hu2024} that demanding evaluation formats hide abilities of smaller models in particular.
Similarly, linear probes on hidden states judge grammaticality better than the model's own string probabilities \cite{wang2026}, and probing reveals linguistic competence that prompting does not \cite{he2025}.
A similar gap between what a model encodes and what it says has been found for truthfulness.
Linear probes and unsupervised methods recover whether a statement is true from hidden states more reliably than the model's own answers \cite{azaria2023,burns2023}.
This study contributes to this line of research with evidence from prosody.
Prompted text LLMs do not demonstrate better performance than lexical baselines in terms of accuracy and provide less contextual information than their representations encode.

\section{Dataset}

\begin{table*}[t]
\caption{The data splits. Words: words with a prominence label (punctuation excluded). Labels: share of words in each 3-way class.}
\label{tab:data}
\centering\small
\begin{tabular}{@{}llrrrrrrl@{}}
\toprule
 & & & & & \multicolumn{3}{c}{Labels (\%)} & \\
\cmidrule(lr){6-8}
Split & LibriTTS subset & Utterances & Words & Speakers & 0 & 1 & 2 & Used for \\
\midrule
\texttt{train\_360} & train-clean-360 & 116,263 & 2,076,289 & 904 & 48.3 & 27.4 & 24.2 & training (text) \\
\texttt{train\_100} & train-clean-100 & 33,042 & 570,592 & 245 & 48.1 & 27.3 & 24.6 & training (speech; matched text) \\
\texttt{dev} & dev-clean & 5,727 & 99,200 & 40 & 47.9 & 27.7 & 24.4 & model, layer and prompt selection \\
\texttt{test} & test-clean & 4,822 & 90,063 & 39 & 48.0 & 27.3 & 24.7 & evaluation \\
\bottomrule
\end{tabular}

\end{table*}

The data are the prominence annotations of Talman et al.~\cite{talman2019}, which give a label to every word of LibriTTS, a corpus of read English derived from LibriSpeech audiobooks \cite{zen2019}.
The labels were produced with the Wavelet Prosody Toolkit \cite{suni2017}, which combines the fundamental frequency, energy and word durations of an utterance into one signal, analyses it with a continuous wavelet transform and assigns each word a score.
The released data store this score next to the label.
We find that the 3-way labels are fixed cut-offs of the score (0 below 0.4, 1 below 1.2, 2 above), without exception over \resval{corpus-words} words, and that the score has been clipped at zero: no value is negative and \resval{corpus-zero-share}\% are exactly zero.
We refer to the classes as 0 (non-prominent), 1 (prominent) and 2 (very prominent); the 2-way task merges 1 and 2.
The annotations follow the LibriTTS splits (Table~\ref{tab:data}), which have no speakers in common.
Text models are trained on \texttt{train\_360}, the larger training set used in 2019.
Since extracting speech representations for every frame of \texttt{train\_360} would be costly, speech models are trained on \texttt{train\_100}, and to separate the effect of the input from that of the training data, we also train two text classifiers on it (Section~\ref{sec:rq3}).
Models, layers and prompts are chosen on the development set, and we report word-level accuracy on the test set, leaving out punctuation.
The three labels have nearly the same distribution in all four splits.
As in 2019, the reported numbers leave out the last test utterance; this changes no result by more than \resval{legacy-full-maxdiff} points.

\textbf{Reproducing the labels.} Since the labels are a deterministic function of the recording and its word alignment, they can in principle be computed again.
The original alignments are not available, so we use the public forced alignments of LibriTTS by Hayashi,\footnote{\url{https://github.com/kan-bayashi/LibriTTSLabel}} made with the Montreal Forced Aligner \cite{mcauliffe2017}, and run the toolkit with the configuration distributed for LibriTTS.
On one test utterance for which the toolkit's own alignment is distributed, both alignments give the original label for \resval{sample-public-same} of \resval{sample-words} words, and the two alignments change the scores by less than 0.05.
On the whole test set, the reconstructed labels agree with the original ones on \resval{toolkit-3}\% (3-way) and \resval{toolkit-2}\% (2-way) of the words, and the scores correlate at $r=\resval{recon-pearson}$.
The remaining differences come from the alignments and from changes to the toolkit since 2019.
The reconstruction serves as the upper reference for speech models: it is the labelling procedure itself, applied to our word boundaries.

\textbf{Readers disagree.} Some texts in LibriTTS are read by several speakers.
On these \resval{agreement-groups} texts, excluding the LibriVox announcements, two readers agree on \resval{agreement-2}\% of the 2-way and \resval{agreement-3}\% of the 3-way labels.
\resval{overlap-utterances} test utterances also occur in the training or development data, read by another speaker.
Every text model is \resval{overlap-diff-min-3}--\resval{overlap-diff-max-3} points less accurate on them than on the rest of the test set (3-way), so having seen the text during training does not help.

\section{Experimental setup}

\begin{table}[t]
\caption{Models, with their Hugging Face identifiers, number of parameters (Whisper and Qwen2-Audio: the whole model) and the revisions used.}
\label{tab:models}
\centering\scriptsize
\resizebox{\columnwidth}{!}{\begin{tabular}{@{}llrl@{}}
\toprule
Model & Hugging Face identifier & Params & Revision \\
\midrule
\multicolumn{4}{@{}l}{\textit{Encoders}} \\
BERT-base & \texttt{google-bert/bert-base-uncased} & 110M & \texttt{86b5e093} \\
BERT-base cased & \texttt{google-bert/bert-base-cased} & 109M & \texttt{cd5ef92a} \\
BERT-large & \texttt{google-bert/bert-large-uncased} & 336M & \texttt{6da4b6a2} \\
ModernBERT-base & \texttt{answerdotai/ModernBERT-base} & 150M & \texttt{8949b909} \\
ModernBERT-large & \texttt{answerdotai/ModernBERT-large} & 396M & \texttt{45bb4654} \\
XLM-R base & \texttt{FacebookAI/xlm-roberta-base} & 279M & \texttt{e73636d4} \\
XLM-R large & \texttt{FacebookAI/xlm-roberta-large} & 561M & \texttt{c23d21b0} \\
\midrule
\multicolumn{4}{@{}l}{\textit{Decoder LLMs}} \\
Qwen3.5 0.8B & \texttt{Qwen/Qwen3.5-0.8B-Base} & 873M & \texttt{dc7cdfe2} \\
Qwen3.5 2B & \texttt{Qwen/Qwen3.5-2B-Base} & 2.3B & \texttt{b1485b2f} \\
Qwen3.5 4B & \texttt{Qwen/Qwen3.5-4B-Base} & 4.7B & \texttt{1001bb4d} \\
Qwen3.5 9B & \texttt{Qwen/Qwen3.5-9B-Base} & 9.7B & \texttt{68c46c4b} \\
Qwen3.8 27B & \texttt{Qwen/Qwen3.8-27B} & 27.8B & \texttt{37561816} \\
OLMo 3 7B & \texttt{allenai/Olmo-3-1025-7B} & 7.3B & \texttt{a81bae42} \\
OLMo 3 32B & \texttt{allenai/Olmo-3-1125-32B} & 32.2B & \texttt{c2b61dae} \\
Muse Glimmer 30B & \texttt{meta-models/Muse-Glimmer-30B} & 29.8B & \texttt{cec94fcf} \\
\midrule
\multicolumn{4}{@{}l}{\textit{Speech}} \\
WavLM-large & \texttt{microsoft/wavlm-large} & 315M & \texttt{c1423ed9} \\
HuBERT-large & \texttt{facebook/hubert-large-ll60k} & 315M & \texttt{ff022d09} \\
Whisper large-v3 (encoder) & \texttt{openai/whisper-large-v3} & 1.5B & \texttt{06f233fe} \\
Qwen2-Audio 7B & \texttt{Qwen/Qwen2-Audio-7B} & 8.4B & \texttt{dd844707} \\
Qwen2-Audio 7B Instruct & \texttt{Qwen/Qwen2-Audio-7B-Instruct} & 8.4B & \texttt{0a095220} \\
\bottomrule
\end{tabular}
}
\end{table}

\textbf{Text models.} The models are listed in Table~\ref{tab:models}.
We fine-tune BERT-base with the original recipe \cite{talman2019}: Adam, learning rate $5\cdot10^{-5}$, batch size 32, two epochs and the final-epoch model, with five seeds.
BERT-large, ModernBERT-large \cite{warner2024} and XLM-R large \cite{conneau2020} are fine-tuned with the same recipe, but with an attention mask and a learning rate of $2\cdot10^{-5}$ (three seeds each).
One of the three 2-way runs of XLM-R large did not train and stayed at the level of the majority class on the development set; we leave such runs out of the averages.
We probe \resval{encoders-probed} frozen encoders and \resval{decoders-probed} decoder LLMs from 0.8B to 30B parameters, representing each word by its first (encoders) or last (decoders) sub-word token.
Decoders only see the left context, so we also give them the sentence twice and read the second copy (\textsc{echo}).
A linear discriminant probe is trained on every layer and the best layer is chosen on the development set.
On that layer we then train three classifiers on the frozen features, logistic regression, a multilayer perceptron and a bidirectional LSTM (BiLSTM), with three seeds each.
As a control, the same BiLSTM is trained on random vectors, one fixed vector per word type, which carry no knowledge from any model.

\textbf{Prompting.} GPT-4o, DeepSeek-V4-Pro and Mistral Large 3 receive the sentence and its numbered words and are asked for one label per word.
We compare three instruction variants, three answer formats and 0, 1, 5 or 10 examples on 500 development utterances, and run the best setting of each model on the test set.
The reasoning model GPT-5.5 is run on 200 test utterances with the setting chosen for GPT-4o.
To see whether more examples help, we give the open model Qwen3.8 27B, which we also probe, between 0 and 500 examples (up to 146k tokens).
It writes its answer word by word and we take the label it finds most probable, so that no answer needs to be parsed.

\textbf{Speech models.} Corpus words are matched to the aligned words, handling expanded abbreviations, unknown words and differences in tokenisation.
We probe every layer of WavLM-large \cite{chen2022}, HuBERT-large \cite{hsu2021}, the Whisper large-v3 encoder \cite{radford2023} and Qwen2-Audio 7B, both its audio encoder and its language model at the positions of the audio, representing each word by the mean of the frames inside its time span.
The speech classifiers are trained on \texttt{train\_100}; to compare text and speech with the same training data, we also train them on BERT-base and Muse Glimmer 30B on \texttt{train\_100}.
Qwen2-Audio 7B Instruct is prompted with the same prompt as above, once with and once without the recording.
As a further reference, a linear classifier uses word timings alone (duration, number of phones, pauses and position in the utterance).

\textbf{Comparing text and speech.} Text and speech models are scored on the same words: those with a time span that starts within the first 30\,s of the utterance, the input window of Whisper and Qwen2-Audio (\resval{common-words} words, \resval{common-share}\% of the test words).
Differences are tested with McNemar's test, and we report utterance-level bootstrap intervals.

\textbf{Evaluation and compute.} As in 2019, the primary metric is word-level accuracy. Due to class imbalance, we also report the macro-F1 scores and recall for each class for the primary systems.
The open models are run on single NVIDIA GH200 GPUs (probing with bfloat16), and the commercial models are run via the Azure API.
The runs used in this paper (including controls not reported here) took a total of \resval{gpu-hours-total} GPU hours. The longest run was prompting Qwen3.8 27B with 500 examples on the test set, taking \resval{gpu-hours-longest} hours.

\section{Results}

\begin{table}[t]
\caption{Test accuracy (\%). Mean $\pm$ standard deviation over seeds where there are several. \textsc{t}: text input, \textsc{a}: audio input. Best layers and models are chosen on the development set. Bold: best text and best speech system in each column (baselines and references excluded). $^{\dagger}$200 test utterances.}
\label{tab:main}
\centering\small
\resizebox{\columnwidth}{!}{\begin{tabular}{@{}llcc@{}}
\toprule
 & System & 2-way & 3-way \\
\midrule
\multicolumn{4}{@{}l}{\textit{Baselines}} \\
\textsc{t} & Majority class & 52.0 & 48.0 \\
\textsc{t} & Word majority (2019) & 80.7 & 62.0 \\
\midrule
\multicolumn{4}{@{}l}{\textit{Fine-tuned}} \\
\textsc{t} & BERT-base (2019 recipe) & 83.0 $\pm$ 0.1 & 68.5 $\pm$ 0.1 \\
\textsc{t} & BERT-large & 83.1 $\pm$ 0.1 & 68.5 $\pm$ 0.1 \\
\textsc{t} & ModernBERT-large & 83.2 $\pm$ 0.0 & \textbf{68.8} $\pm$ 0.1 \\
\textsc{t} & XLM-R large & \textbf{83.3} $\pm$ 0.1 & \textbf{68.8} $\pm$ 0.2 \\
\midrule
\multicolumn{4}{@{}l}{\textit{Frozen, linear probe}} \\
\textsc{t} & BERT-base & 81.8 & 64.6 \\
\textsc{t} & best text: Muse Glimmer 30B, echo & 82.6 & 66.9 \\
\textsc{a} & best audio: Whisper large-v3 (encoder) & 88.2 & 78.7 \\
\midrule
\multicolumn{4}{@{}l}{\textit{Frozen, BiLSTM classifier}} \\
\textsc{t} & best text (3-way): Qwen3.5 0.8B, echo & 82.9 $\pm$ 0.0 & 68.0 $\pm$ 0.1 \\
\textsc{t} & best text (2-way): ModernBERT-large & 83.0 $\pm$ 0.0 & 68.0 $\pm$ 0.1 \\
\textsc{a} & best audio: Whisper large-v3 (encoder) & \textbf{91.0} $\pm$ 0.2 & \textbf{84.7} $\pm$ 0.2 \\
\midrule
\multicolumn{4}{@{}l}{\textit{Prompting}} \\
\textsc{t} & GPT-4o & 79.7 & 59.4 \\
\textsc{t} & DeepSeek-V4-Pro & 79.2 & 60.5 \\
\textsc{t} & Mistral Large 3 & 78.9 & 59.7 \\
\textsc{t} & GPT-5.5$^{\dagger}$ & 78.9 & 58.8 \\
\textsc{t} & Qwen3.8 27B, 500 examples & 78.3 & 62.0 \\
\textsc{a} & Qwen2-Audio, with recording & 51.0 & 39.2 \\
\textsc{t} & Qwen2-Audio, text only & 51.6 & 43.1 \\
\midrule
\multicolumn{4}{@{}l}{\textit{Audio references}} \\
\textsc{a} & Duration only & 83.2 & 65.0 \\
\textsc{a} & Labelling procedure & 94.9 & 91.4 \\
\bottomrule
\end{tabular}
}
\end{table}

Table~\ref{tab:main} gives an overview.
Our reproduction of the 2019 model reaches \resval{repro-2}\,$\pm$\,\resval{repro-sd-2}\% (2-way) and \resval{repro-3}\,$\pm$\,\resval{repro-sd-3}\% (3-way), within 0.2 points of the published \resval{paper2019-2}\% and \resval{paper2019-3}\%.
A baseline that gives each word its most frequent label in the training data already reaches \resval{wordmaj-2}\% and \resval{wordmaj-3}\%.

\subsection{Text models}

If more text and more parameters taught models more about prominence, the largest LLMs should come out ahead.
Table~\ref{tab:main} shows that they barely do.
The linear probe of the best LLM, \resval{lin-text-model-3}, reaches \resval{lin-text-2}\% and \resval{lin-text-3}\%.
Scaling the Qwen models from 0.8B to 27B parameters raises the 3-way accuracy of the linear probe only from \resval{size-0.8-3}\% to \resval{size-27-3}\%, and that of the BiLSTM classifier not at all.
With a BiLSTM classifier, the best frozen model reaches \resval{bilstm-text-2}\% and \resval{bilstm-text-3}\%.
Fine-tuning larger and newer encoders does not move the ceiling either: BERT-large, ModernBERT-large and XLM-R large reach \resval{ft-bert-large-uncased-3}\%, \resval{ft-modernbert-large-3}\% and \resval{ft-xlm-roberta-large-3}\% on the 3-way task, against \resval{repro-3}\% for BERT-base.

Increasing the amount of training data does not raise the performance ceiling.
The linear probe on BERT-base reaches \resval{lc-linear-5000}\% with 5,000 training utterances, and \resval{lc-linear-116263}\% when using all \resval{data-train360-utts}. Additionally, fine-tuning on 1/10 of \texttt{train\_360} yields \resval{lc-finetune-11626}\%, whereas using all of it yields \resval{lc-finetune-116263}\%.
Therefore, what a text model can learn about prominence can be learned even with a small amount of data.
Since even BiLSTM classifiers on random word vectors without a pre-trained model exhibit \resval{ctl-random-bilstm-3}\% performance, most of the performance achieved by text models can be learned solely from word identities and labels.
Furthermore, although Talman et al.~\cite{talman2019} suggested it could help, the preceding discourse does not help. Providing BERT-base with one or three preceding sentences of a given chapter, or ModernBERT-large with three preceding sentences, changes the 3-way accuracy of both the probes and the classifiers by at most \resval{ctx-max-gain-3} points.

Providing the entire sentence (\textsc{echo}) to the decoders adds \resval{echo-gain-min-3}--\resval{echo-gain-max-3} points on the 3-way task, suggesting that the following words matter, as expected from the accent at the end of a phrase.

The models also make the same mistakes.
Among the \resval{consensus-systems} text systems (linear probes and classifiers for all text models and conditions, and therefore not independent systems), any two agree with each other on average for \resval{consensus-mutual-2}\% of the words (2-way). However, each system agrees with the gold labels on only \resval{consensus-acc-2}\%.
For \resval{consensus-unanimous-wrong-2}\% of the test words, all systems select the same incorrect label, which accounts for \resval{consensus-shared-errors-2}\% of the average system's 2-way errors (\resval{consensus-shared-errors-3}\% for 3-way).
Therefore, while models with different architectures, sizes, and training data predict the same reading of the text, the gold labels differ from the predictions for the same words.
This can be seen as an indicator that the ceiling is determined by the task, namely, how much of prominence is left to the reader.

\begin{figure}[t]
\centering
\resizebox{0.60\columnwidth}{!}{\input{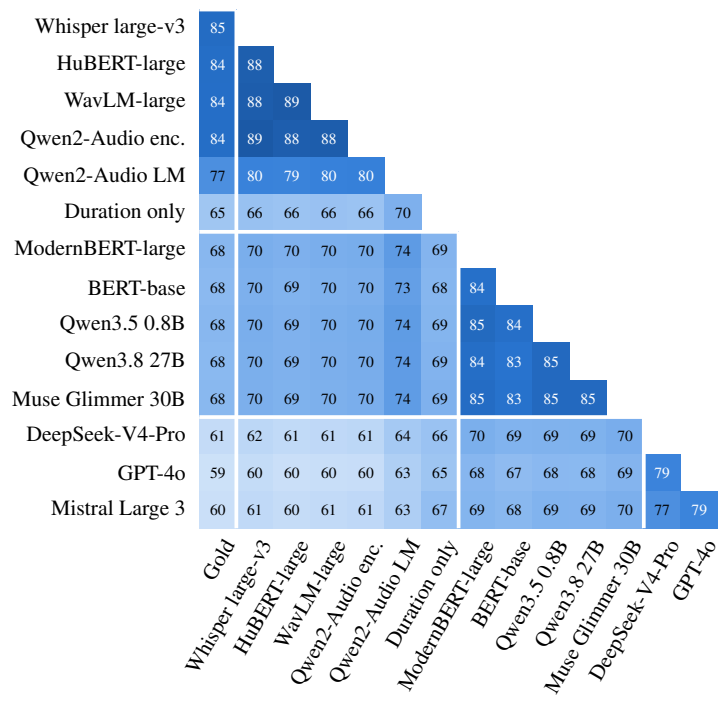}}
\caption{How often two systems give a word the same 3-way label (\%), on the \resval{vis-words} test words used for the text and speech comparison. Text and speech models are BiLSTM classifiers; white lines separate the gold labels, speech models, text models and prompted LLMs.}
\label{fig:agreement}
\end{figure}

Figure~\ref{fig:agreement} shows the same pattern for all systems on the words used in Section~\ref{sec:rq3}.
The text classifiers agree with each other on \resval{agr-text-text}\% of the words on average and with the gold labels on only \resval{agr-gold-text}\%.
The speech models agree with each other about as often (\resval{agr-speech-speech}\%), but with the gold labels on \resval{agr-gold-speech}\%.
The two kinds of model thus form two groups, which agree with each other on only \resval{agr-text-speech}\% of the words.

\begin{figure*}[t]
\centering
\begin{tikzpicture}
\begin{groupplot}[group style={group size=2 by 1, horizontal sep=1.6cm}, width=0.47\textwidth, height=3.2cm,
  xlabel={Relative layer depth}, xmin=0, xmax=1,
  legend style={font=\scriptsize, at={(0.98,0.03)}, anchor=south east, draw=none, fill=white, fill opacity=0.8, text opacity=1},
  legend cell align=left, tick label style={font=\scriptsize}, label style={font=\scriptsize},
  title style={font=\small}]
\nextgroupplot[title={Text (linear probe, \texttt{train\_360})}, ylabel={3-way test accuracy (\%)}, ymin=50, ymax=70]
\addplot+[mark=none, thick] table[x=depth, y=acc3, col sep=comma] {results/layers_bert_base_uncased.csv};
\addlegendentry{BERT-base}
\addplot+[mark=none, thick] table[x=depth, y=acc3, col sep=comma] {results/layers_muse_glimmer_30b.csv};
\addlegendentry{Muse Glimmer 30B, echo}
\addplot+[mark=none, thick] table[x=depth, y=acc3, col sep=comma] {results/layers_qwen3_8_27b.csv};
\addlegendentry{Qwen3.8 27B, echo}
\nextgroupplot[title={Speech (linear probe, \texttt{train\_100})}, ymin=40, ymax=82, legend columns=2, legend style={font=\tiny, /tikz/every even column/.append style={column sep=3pt}}]
\addplot+[mark=none, thick] table[x=depth, y=acc3, col sep=comma] {results/layers_wavlm_large.csv};
\addlegendentry{WavLM-large}
\addplot+[mark=none, thick] table[x=depth, y=acc3, col sep=comma] {results/layers_hubert_large.csv};
\addlegendentry{HuBERT-large}
\addplot+[mark=none, thick] table[x=depth, y=acc3, col sep=comma] {results/layers_whisper_large_v3.csv};
\addlegendentry{Whisper large-v3 encoder}
\addplot+[mark=none, thick] table[x=depth, y=acc3, col sep=comma] {results/layers_qwen2_audio_7b_encoder.csv};
\addlegendentry{Qwen2-Audio encoder}
\addplot+[mark=none, thick, dashed] table[x=depth, y=acc3, col sep=comma] {results/layers_qwen2_audio_7b.csv};
\addlegendentry{Qwen2-Audio language model}
\end{groupplot}
\end{tikzpicture}
\caption{3-way test accuracy of a linear probe on each layer, by relative depth (0: embeddings, 1: last layer).}
\label{fig:layers}
\end{figure*}
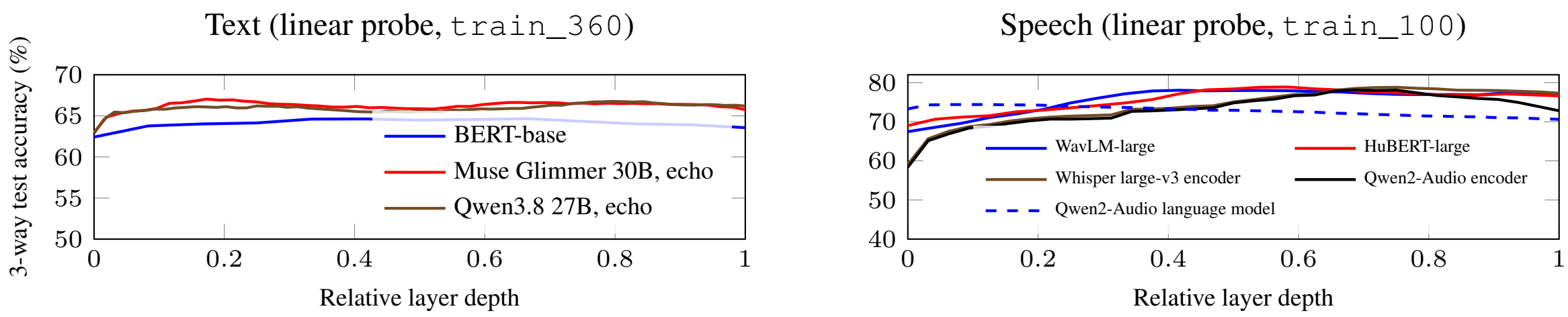

\subsection{Probing and prompting}

\begin{table}[t]
\caption{The same model probed and prompted (test accuracy, \%). The linear probes are trained on the same data at every layer; the embedding layer encodes each word without its context. \textsc{t}: text input, \textsc{a}: audio input. Bold: best access for each model.}
\label{tab:paired}
\centering\small
\resizebox{\columnwidth}{!}{\begin{tabular}{@{}llcc@{}}
\toprule
Model & Access & 2-way & 3-way \\
\midrule
Qwen3.8 27B (\textsc{t}) & linear probe, embedding layer & 81.2 & 62.9 \\
 & linear probe, best layer & \textbf{82.3} & \textbf{66.7} \\
 & prompted, no examples & 78.8 & 58.6 \\
 & prompted, 500 examples & 78.3 & 62.0 \\
\midrule
Qwen2-Audio 7B (\textsc{a}) & linear probe (LM) & \textbf{86.4} & \textbf{74.4} \\
 & prompted (Instruct) & 51.0 & 39.2 \\
\bottomrule
\end{tabular}
}
\end{table}

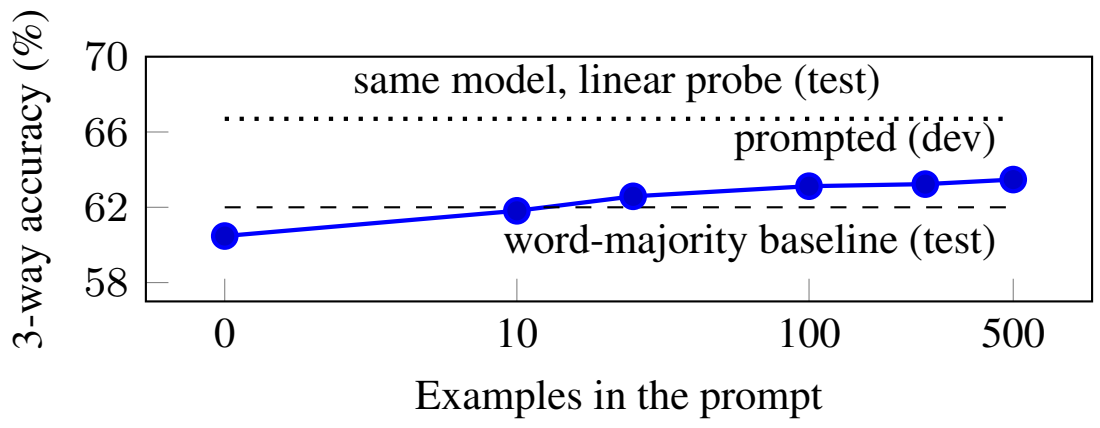
\begin{figure}[t]
\centering
\begin{tikzpicture}
\begin{axis}[width=0.98\columnwidth, height=3.2cm, xmode=log, log basis x=10,
  xlabel={Examples in the prompt}, ylabel={3-way accuracy (\%)},
  xtick={1,10,100,500}, xticklabels={0,10,100,500}, ymin=57, ymax=70, ytick={58,62,66,70},
  xtick pos=bottom, ytick pos=left,
  tick label style={font=\footnotesize}, label style={font=\footnotesize}]
\addplot+[mark=*, thick, mark size=2.2pt] table[x=shots_plot, y=acc3, col sep=comma] {results/icl_curve.csv};
\addplot[dotted, thick, domain=1:500] {\resval{p-qwen-lin-3}};
\addplot[dashed, domain=1:500] {\resval{wordmaj-3}};
\node[anchor=south, font=\footnotesize] at (axis cs:22,\resval{p-qwen-lin-3}) {same model, linear probe (test)};
\node[anchor=south east, font=\footnotesize] at (axis cs:500,63.6) {prompted (dev)};
\node[anchor=north east, font=\footnotesize] at (axis cs:500,\resval{wordmaj-3}) {word-majority baseline (test)};
\end{axis}
\end{tikzpicture}
\caption{Prompting Qwen3.8 27B with more examples (500 development utterances).}
\label{fig:icl}
\end{figure}

Since the contextual representations of LLMs contain information about prominence beyond word identity, one might expect the models to identify prominent words when asked.
However, this is not the case.
The best commercial models demonstrated an accuracy of \resval{prompt-best-3}\% on the 3-way task (\resval{prompt-best-model-3}) and \resval{prompt-best-2}\% on the 2-way task (\resval{prompt-best-model-2}), which is lower than the word-majority baseline (\resval{wordmaj-3}\% and \resval{wordmaj-2}\%).
The format of the answer matters more than the wording of the instructions, and examples help little: 1 to 10 examples add only about 1 point on the 3-way task and nothing on the 2-way task.
The reasoning model GPT-5.5 is not better either, but since it was run on only 200 utterances, its result is for reference only.

Is information missing from the prompted model, or is the model unable to report that information?
Table~\ref{tab:paired} compares the same model accessed in two ways.
When probing with a linear classifier, Qwen3.8 27B achieved a performance of \resval{p-qwen-lin-3}\% in the 3-way task. When prompted with 500 examples, it achieved \resval{icl-test-3}\%, matching the performance of the word-majority baseline, while in the 2-way task, it achieved \resval{icl-test-2}\%, showing lower performance than the baseline.
The linear probe can do little on its own. In the model's embedding layer, which processes each word without context, it achieved \resval{emb-qwen-3}\%, showing performance close to the baseline.
Therefore, the information the probe finds beyond the word's identity lies in the model's contextual representations, and the prompted model does not report this information.
Here, \emph{represented} simply means that a linear classifier trained on labelled data can recover information from the hidden states. This does not imply that the model possesses a concept of prominence or can use it during generation.

The probe and the prompt use labelled data in different ways, and the comparison should be read with this in mind.
The probe is a measuring instrument: it is trained on labelled words to read out what the frozen hidden states already encode, and training it changes nothing in the model.
The examples in the prompt, in contrast, are input to the model, which has to make use of them itself.
The data the probe is trained on is therefore not supervision for the language model, and the number of utterances the probe sees is not an amount of help the model has received.
What the probe's training data does affect is how precisely the instrument reads out the representations, and a probe can also learn part of the task on its own \cite{hewitt2019}.
Training the probe on only as many utterances as the prompt holds would therefore make the instrument less precise, not the comparison fairer; the control for what the probe learns on its own is the same probe on the embedding layer.
For this reason we judge what the representations contain by comparing the same probe, trained on the same data, on different representations: the contextual layer of Qwen3.8 27B against its embedding layer (Table~\ref{tab:paired}).
BiLSTM classifiers are less suitable for comparison because they can learn a large portion of the task on their own. For random word vectors, BiLSTM classifiers achieve a performance of \resval{ctl-random-bilstm-3}\%, which is close to \resval{p-qwen-bilstm-3}\% achieved on Qwen3.8 27B (\resval{ctl-qwen-layer0-bilstm-3}\% on its embedding layer).
In the development subset, examples help only slowly, from \resval{icl-dev-0-3}\% without examples to \resval{icl-dev-500-3}\% when 500 examples are added, and they are of no help at all in the 2-way task (Figure~\ref{fig:icl}).

The speech-text LLM exhibits the same pattern in a stronger form.
On Qwen2-Audio's audio encoder, the BiLSTM classifier achieves \resval{p-qaenc-bilstm-3}\%, and on its language model it achieves \resval{p-qalm-bilstm-3}\%, demonstrating far higher performance than all text models. However, when prompted with the recording, the instruction-tuned model only achieves \resval{ap-audio-3}\% (when used without the recording, \resval{ap-text-3}\%), showing lower performance than the majority class.
Allowed to answer freely on \resval{fg-audio-n} test sentences, it mostly repeats the numbered words without labels (\resval{fg-audio-enumeration} sentences with the recording), and where it does give labels, it marks almost every word as non-prominent (\resval{fg-audio-zero} of \resval{fg-audio-labelled} words).
Therefore, the instruction-tuned model fails to perform the task properly, whether it responds freely or in the required format, even though its encoder represents prominence better than any text model.

\begin{table}[t]
\caption{How each system labels the test set (3-way): macro-F1 and recall of each class, and the share of content words, function words and utterance-final words given the highest level (2). \textsc{t}: text classifier, \textsc{p}: prompted, \textsc{a}: speech classifier.}
\label{tab:behaviour}
\centering\small
\resizebox{\columnwidth}{!}{\begin{tabular}{@{}llrrrrrrr@{}}
\toprule
 & & & \multicolumn{3}{c}{Recall (\%)} & \multicolumn{3}{c}{Labelled 2 (\%)} \\
\cmidrule(lr){4-6}\cmidrule(l){7-9}
 & System & Macro-F1 & 0 & 1 & 2 & content & function & last \\
\midrule
 & Gold labels & -- & -- & -- & -- & 42.2 & 9.2 & 31.9 \\
\textsc{t} & ModernBERT-large, BiLSTM classifier & 63.9 & 83.5 & 45.2 & 63.1 & 46.7 & 6.7 & 35.4 \\
\textsc{t} & Qwen3.8 27B, BiLSTM classifier & 63.8 & 83.2 & 48.4 & 59.2 & 43.1 & 6.0 & 31.7 \\
\textsc{p} & Qwen3.8 27B, prompted (500 ex.) & 55.9 & 84.2 & 35.4 & 48.3 & 44.2 & 3.4 & 45.5 \\
\textsc{p} & DeepSeek-V4-Pro, prompted (5 ex.) & 54.7 & 80.0 & 31.1 & 55.1 & 56.8 & 4.8 & 77.1 \\
\textsc{p} & GPT-4o, prompted (5 ex.) & 54.9 & 75.4 & 36.7 & 53.5 & 58.7 & 2.8 & 81.4 \\
\textsc{a} & Whisper large-v3, BiLSTM classifier & 83.0 & 90.4 & 72.4 & 86.6 & 43.4 & 9.3 & 32.7 \\
\bottomrule
\end{tabular}
}
\end{table}

Table~\ref{tab:behaviour} shows what the prompted models say instead: how often each system gives the highest level to content words, function words and the last word of the utterance.
The BiLSTM classifier on Qwen3.8 27B reproduces the proportions of the gold labels almost exactly, while the prompted models give the highest level to the last word far too often: \resval{pp-gpt4o-last-top}\% of the time for GPT-4o and \resval{pp-deepseek-last-top}\% for DeepSeek-V4-Pro, against \resval{pp-gold-last-top}\% in the gold labels.
This is not an effect of our instructions, which mention that prominence tends to fall towards the end of a phrase.
With the minimal instructions, which say nothing about position, the three commercial models still give the highest level to \resval{pp-dev-minimal-last-top-min}--\resval{pp-dev-minimal-last-top-max}\% of the last words on the development subset (gold: \resval{pp-dev-gold-last-top}\%).
With 500 examples, Qwen3.8 27B corrects much of this (\resval{pp-qwen-icl-last-top}\%), but it still falls \resval{pp-qwen-icl-content-gap} points behind the classifier on its representations on content words and \resval{pp-qwen-icl-function-gap} points on function words.
The prompted models thus seem to answer with a rule of thumb about where the main accent of a sentence goes, while a classifier trained on the same model's representations comes closer to how the sentences were actually read.
Most of the difference is on content words, that is, on deciding which of the words that could be prominent actually are.

Accuracy hides some of the differences, as about half of the words are non-prominent and all systems classify most of the words correctly.
Table~\ref{tab:behaviour} shows the macro-F1 scores and recall for each class.
The classifier and the prompted model built on Qwen3.8 27B show a \resval{icl-probe-f1-gap}-point difference in macro-F1, larger than in accuracy.
This difference appears in the prominent classes. The classifier detects \resval{cls-qwen-recall-1}\% of words labelled 1 and \resval{cls-qwen-recall-2}\% of words labelled 2, whereas the prompted model detects \resval{cls-qwen-icl-recall-1}\% and \resval{cls-qwen-icl-recall-2}\%.
When asked directly, the model misclassifies \resval{cls-qwen-icl-conf-20}\% of very prominent words as non-prominent, whereas the classifier misclassifies \resval{cls-qwen-conf-20}\%. Therefore, prompts not only blur the boundaries between levels but also miss prominent words entirely.
On the other hand, the speech model rarely confuses the two extremes (\resval{cls-whisper-conf-20}\% of words labelled 2 are predicted as 0), and most errors occur between adjacent levels.

\begin{table}[t]
\caption{How often each system gives a word the label that the word most often has in the training data (the word-majority lookup), and, on the words where it does not, how often the system and the lookup are right (\%, 3-way, test set).}
\label{tab:lexical}
\centering\small
\resizebox{\columnwidth}{!}{\begin{tabular}{@{}llrrrr@{}}
\toprule
 & & & & \multicolumn{2}{c}{Right where it departs} \\
\cmidrule(l){5-6}
 & System & Same as lookup & Departs & system & lookup \\
\midrule
\textsc{t} & ModernBERT-large, BiLSTM classifier & 71.7 & 28.3 & 51.8 & 30.7 \\
\textsc{t} & Qwen3.8 27B, BiLSTM classifier & 72.4 & 27.6 & 51.4 & 30.8 \\
\textsc{p} & Qwen3.8 27B, prompted (500 ex.) & 69.3 & 30.7 & 39.5 & 39.8 \\
\textsc{p} & DeepSeek-V4-Pro, prompted (5 ex.) & 68.8 & 31.2 & 37.1 & 42.1 \\
\textsc{p} & GPT-4o, prompted (5 ex.) & 68.0 & 32.0 & 36.0 & 44.2 \\
\bottomrule
\end{tabular}
}
\end{table}

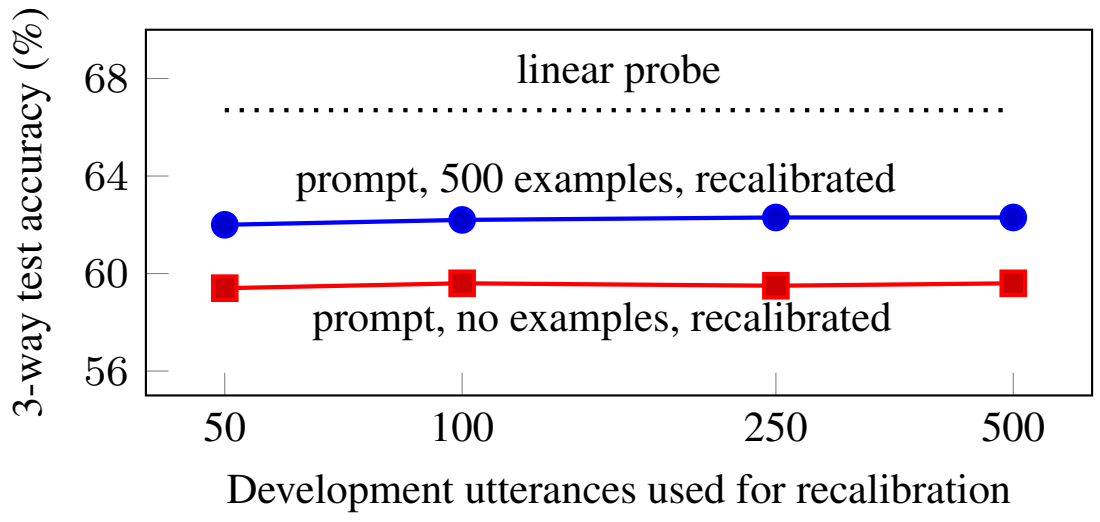
\begin{figure}[t]
\centering
\begin{tikzpicture}
\begin{axis}[width=0.98\columnwidth, height=4.0cm, xmode=log, log basis x=10,
  xlabel={Development utterances used for recalibration}, ylabel={3-way test accuracy (\%)},
  xtick={50,100,250,500}, xticklabels={50,100,250,500}, ymin=55, ymax=70, ytick={56,60,64,68}, xtick pos=bottom, ytick pos=left,
  tick label style={font=\footnotesize}, label style={font=\footnotesize}]
\addplot+[mark=*, thick, mark size=2.2pt] coordinates {(50,\resval{v2-sat-k500-3-n50}) (100,\resval{v2-sat-k500-3-n100}) (250,\resval{v2-sat-k500-3-n250}) (500,\resval{v2-sat-k500-3-n500})} node[pos=0.5, above, font=\footnotesize, text=black] {prompt, 500 examples, recalibrated};
\addplot+[mark=square*, thick, mark size=2.2pt] coordinates {(50,\resval{v2-sat-k0-3-n50}) (100,\resval{v2-sat-k0-3-n100}) (250,\resval{v2-sat-k0-3-n250}) (500,\resval{v2-sat-k0-3-n500})} node[pos=0.5, below, font=\footnotesize, text=black] {prompt, no examples, recalibrated};
\addplot[dotted, thick, domain=50:500] {\resval{p-qwen-lin-3}} node[pos=0.5, above, font=\footnotesize] {linear probe};
\end{axis}
\end{tikzpicture}
\caption{Qwen3.8 27B: its own label probabilities recalibrated on 50 to 500 development utterances (mean of five random subsets below 500); test set.}
\label{fig:recalibration}
\end{figure}

Are the prompted models simply wrong about where the cut-offs between the levels lie?
If so, recalibrating their own label probabilities should close part of the gap.
We fit a logistic map from the three log-probabilities to the labels on the 500 development utterances and apply it to the test set.
For Qwen3.8 27B with 500 examples, this moves the 3-way accuracy only from \resval{v2-sat-k500-3-own}\% to \resval{cal-3way-k500-logistic}\%, and the result is the same whether the map is fitted on 50, 100, 250 or 500 utterances (\resval{v2-sat-k500-3-n50}--\resval{v2-sat-k500-3-n500}\%; Figure~\ref{fig:recalibration}).
More labelled data thus changes the recalibrated answer little, and it stays well below the \resval{p-qwen-lin-3}\% of the probe.
Comparing the probe with the prompt is therefore not a comparison of two learners given different amounts of data: the probe measures what the representations encode, and the prompt asks the model to report it.

We then ask whether the answer carries any information about the context at all.
We take the word types (lower-cased) that occur at least ten times in the test set and are prominent in some of their occurrences and not in others (\resval{v2-wt-types} types, \resval{v2-wt-share}\% of the test words).
Within each type, a system that only knows which word it is cannot tell the occurrences apart, and we measure how well a system's probability of prominence does so with the area under the ROC curve (AUC; ties count as half), averaged over the types.
The prompted model with 500 examples reaches \resval{v2-wt-p500}, against \resval{v2-wt-p0} without examples and \resval{v2-wt-pos} for a classifier that sees only the position of the word in the sentence (\resval{v2-wt-pos-nb} against \resval{v2-wt-p500-nb} for the prompted model when final words and words before punctuation are left out, a difference of \resval{v2-wt-p500-minus-pos-nb} with 95\% interval \resval{v2-wt-p500-minus-pos-nb-cilo}--\resval{v2-wt-p500-minus-pos-nb-cihi}).
The answer is therefore not blind to context, and the examples add to it (\resval{v2-wt-gain} [\resval{v2-wt-gain-cilo}, \resval{v2-wt-gain-cihi}]), also when final words and words before punctuation are left out (\resval{v2-wt-gain-nb}, over the \resval{v2-wt-types-nb} types that then qualify).
It carries less contextual information than the representations, however.
The probe on all the training data reaches \resval{v2-wt-probe}, \resval{v2-wt-probe-minus-p500} more than the recalibrated answer (95\% interval \resval{v2-wt-probe-minus-p500-cilo}--\resval{v2-wt-probe-minus-p500-cihi}, resampling word types), and the probe on the embedding layer, which sees no context, \resval{v2-wt-emb}.
Measured from that floor, the model's own answer covers about \resval{v2-wt-share-of-probe}\% of the distance to the probe, and the gap remains when final words and words before punctuation are left out (\resval{v2-wt-probe-minus-p500-nb}).

To see whether the model reports more when asked differently, we tried two other ways of asking Qwen3.8 27B on the development subset.
Asking which words would be read longest and loudest, instead of which are prominent, gives \resval{v2-e2b-ac500-3}\% with 500 examples against \resval{v2-e2b-ac500-3-ref}\% for our instruction, and \resval{v2-e2b-ac0-3}\% against \resval{v2-e2b-ac0-3-ref}\% without examples.
Asking for every pair of words which of the two is more emphasised, in both orders, and turning each word's mean win probability into three levels with cut-offs fitted on one half of the development utterances and applied to the other, gives \resval{v2-e2b-pw-3}\%, against \resval{v2-e2b-pw-3-ref}\% for our instruction scored the same way.
Neither beats the model's own labels.

Table~\ref{tab:lexical} looks at the same question from the side of the lexical baseline.
All text systems give most words the label they most often have in the training data, the classifiers on \resval{lex-qwen-agree}\% of the words and the prompted models on \resval{lex-gpt4o-agree}--\resval{lex-qwen-icl-agree}\%.
The difference is in what happens when they depart from it.
Where the classifier on Qwen3.8 27B departs from the lookup, it is right on \resval{lex-qwen-dep_right}\% of the words and the lookup only on \resval{lex-qwen-dep_wm_right}\%, so the classifier uses the context to correct the lexical default.
Where the prompted models depart from it, they gain nothing in accuracy: the lookup would have been right about as often as the model, or more often (e.g., \resval{lex-qwen-icl-dep_wm_right}\% against \resval{lex-qwen-icl-dep_right}\% for Qwen3.8 27B with 500 examples, and \resval{lex-gpt4o-dep_wm_right}\% against \resval{lex-gpt4o-dep_right}\% for GPT-4o).
These departures are not random, however, since the lookup is right on them far less often than on the test set as a whole (\resval{wordmaj-3}\%): the prompted model departs from the lexical default where that default tends to be wrong (the contextual information measured above), but it does not choose the right alternative often enough to gain accuracy.

The probe's advantage over the prompt is larger on the last word of the utterance (\resval{v2-e2a-gap-final} points) than on words inside the sentence (\resval{v2-e2a-gap-other} points), which fits a probe that learns part of the lengthening at the end of a phrase that the labels include.
However, final words and words before punctuation make up \resval{v2-e2a-boundary-word-share}\% of the words and carry only \resval{v2-e2a-boundary-gap-share}\% of the gap, so this is one part of the explanation, not most of it.

\subsection{Speech models}
\label{sec:rq3}

\begin{table}[t]
\caption{Accuracy (\%) on the same \resval{common-words} test words, best systems of each kind. All differences from the best text model are significant (McNemar, $p<0.05$) unless marked $^{\circ}$. Bold: best text and best speech system (references excluded).}
\label{tab:gap}
\centering\small
\resizebox{\columnwidth}{!}{\begin{tabular}{@{}llccc@{}}
\toprule
 & System & 2-way & 3-way & 95\% CI (3-way) \\
\midrule
\textsc{a} & Labelling procedure & 95.0 & 91.5 & 91.3--91.7 \\
\textsc{a} & Whisper large-v3 (encoder), BiLSTM classifier & \textbf{90.9} & \textbf{84.6} & 84.3--84.9 \\
\textsc{a} & Whisper large-v3 (encoder), linear probe & 88.3 & 78.8 & 78.5--79.1 \\
\textsc{a} & Qwen2-Audio 7B (LM), BiLSTM classifier & 87.4 & 76.7 & 76.4--77.0 \\
\textsc{a} & Duration only & 83.2$^{\circ}$ & 65.0 & 64.7--65.4 \\
\textsc{t} & Qwen3.5 0.8B, echo, BiLSTM classifier & 82.9$^{\circ}$ & \textbf{68.1} & 67.7--68.4 \\
\textsc{t} & ModernBERT-large, BiLSTM classifier & \textbf{83.1} & \textbf{68.1}$^{\circ}$ & 67.7--68.4 \\
\textsc{t} & Muse Glimmer 30B, echo, BiLSTM classifier, train\_100 & 82.8 & 67.5 & 67.2--67.9 \\
\textsc{t} & Muse Glimmer 30B, echo, linear probe & 82.6 & 66.9 & 66.6--67.2 \\
\textsc{t} & Qwen3.8 27B, prompted & 78.3 & 62.0 & 61.7--62.4 \\
\textsc{a} & Qwen2-Audio 7B Instruct, prompted & 51.0 & 39.2 & 38.8--39.7 \\
\bottomrule
\end{tabular}
}
\end{table}

For the same words, the best speech model, the Whisper encoder-based BiLSTM classifier, achieved a performance of \resval{common-audio-3}\% in the 3-way task and \resval{common-audio-2}\% in the 2-way task.
The best text models achieved performances of \resval{common-text-3}\% and \resval{common-text-2}\% (Table~\ref{tab:gap}; all systems used in this comparison were selected on the development dataset).
Therefore, the recordings add \resval{gap-3} points in the 3-way task and \resval{gap-2} points in the 2-way task.
Since the labels were calculated from the same recordings, some of these benefits are inherent in the task itself (Section~\ref{sec:discussion}).
These differences do not stem from the training data. When using the same \texttt{train\_100} data, the BiLSTM classifiers for BERT-base and Muse Glimmer 30B achieved performances of \resval{p-bert100-bilstm-3}\% and \resval{p-glimmer100-bilstm-3}\% (3-way), respectively, while Whisper reached \resval{p-whisper-bilstm-3}\%.
All four speech models ultimately showed similar performance to each other (for the BiLSTM classifier, \resval{p-wavlm-bilstm-3}--\resval{p-whisper-bilstm-3}\%), and the linear probes peaked in the middle and upper layers, specifically at a depth between \resval{p-wavlm-depth}\% and \resval{p-whisper-depth}\% (Figure~\ref{fig:layers}).
Compared to the best text models overall, ModernBERT-large and XLM-R large (both \resval{text-best-3}\%), the best speech model is ahead by \resval{gap-best-3} points across the entire test set.

How much of the gap is timing alone?
A classifier over word durations, phones and pauses reaches \resval{duration-2}\% on the 2-way task, as much as the best text model, but only \resval{duration-3}\% on the 3-way task, below it.
Duration thus tells whether a word is prominent about as well as the text does, but not how prominent it is; for that, the models need pitch and loudness as well.
Running the labelling procedure itself on our alignments gives \resval{toolkit-3}\% and \resval{toolkit-2}\%, which leaves \resval{toolkit-gap-3} points (3-way) for speech models to learn, part of which is noise from the alignments.

The speech-text LLM loses some of what its encoder has learned.
Within the encoder, the linear probe peaks at layer \resval{p-qaenc-layer} and declines towards the encoder's output; the language model receives this weaker signal, keeps it for a few layers and then slowly loses it (Figure~\ref{fig:layers}).
The best layer of the language model, \resval{p-qalm-layer}, is near its input, and the BiLSTM classifier on it stays \resval{qa-loss-3} points below that on the encoder.

\section{Analysis}

\subsection{Word types}

\begin{table}[t]
\caption{3-way accuracy (\%) of the best text and the best speech model on the same words, by word type and by whether the text models agree. $\Delta$: speech minus text.}
\label{tab:breakdown}
\centering\small
\resizebox{\columnwidth}{!}{\begin{tabular}{@{}lrccc@{}}
\toprule
Words & $n$ & Text & Audio & $\Delta$ \\
\midrule
content words & 42,281 & 56.1 & 79.0 & +22.9 \\
function words & 47,654 & 78.7 & 89.6 & +10.8 \\
first word & 4,740 & 67.0 & 83.4 & +16.4 \\
medial & 80,394 & 69.1 & 85.2 & +16.1 \\
last word & 4,801 & 51.7 & 75.1 & +23.4 \\
new (content) & 29,629 & 56.4 & 79.1 & +22.7 \\
given (content) & 12,652 & 55.3 & 78.8 & +23.5 \\
unseen in training & 1,341 & 60.3 & 77.3 & +16.9 \\
seen 1--10 times & 4,385 & 58.9 & 80.1 & +21.2 \\
seen $>$1,000 times & 53,641 & 75.9 & 88.4 & +12.5 \\
every text model wrong & 9,699 & 0.0 & 58.9 & +58.9 \\
text models disagree & 44,625 & 57.4 & 80.2 & +22.8 \\
every text model right & 35,611 & 100.0 & 97.1 & -2.9 \\
\midrule
all common words & 89,935 & 68.1 & 84.6 & +16.5 \\
\bottomrule
\end{tabular}
}
\end{table}

To gain a better understanding of where the recording helps, we break the comparison down by word type (Table~\ref{tab:breakdown}).
The gain is about twice as large for content words (\resval{bd-3-content-diff} points) as for function words (\resval{bd-3-function-diff} points), and it is largest on the last word of the utterance (\resval{bd-3-final-diff} points).
Given and new content words gain about equally (\resval{bd-3-given-diff} and \resval{bd-3-new-diff} points), which suggests that the speech models do not simply learn information structure better than the text models.
Most importantly, on the words on which every text model makes the same mistake, the speech model is right on \resval{bd-3-all-wrong-audio}\% of them, while it keeps \resval{bd-3-all-right-audio}\% of the words that every text model gets right.
The words that text cannot predict are thus largely the words where the recording decides.

\subsection{Readers of the same text}

\begin{table}[t]
\caption{3-way agreement (\%) of each system's predictions with the reader of the test utterance (own) and with another speaker reading the same text in the development set (other), on \resval{rt-words} words in \resval{rt-utterances} test utterances. Classifiers are BiLSTMs. \textsc{a}: speech, \textsc{t}: text, \textsc{p}: prompted. $\Delta$: own minus other.}
\label{tab:readers}
\centering\small
\resizebox{\columnwidth}{!}{\begin{tabular}{@{}llrrr@{}}
\toprule
 & System & Own reader & Other reader & $\Delta$ \\
\midrule
\textsc{a} & Whisper large-v3 & 80.3 & 65.8 & +14.5 \\
\textsc{a} & WavLM-large & 81.3 & 65.5 & +15.8 \\
\textsc{t} & ModernBERT-large & 67.4 & 70.6 & -3.2 \\
\textsc{t} & Qwen3.8 27B & 65.7 & 69.9 & -4.2 \\
\textsc{p} & Qwen3.8 27B, 500 examples & 60.8 & 62.0 & -1.2 \\
\textsc{p} & DeepSeek-V4-Pro, 5 examples & 59.1 & 58.7 & +0.4 \\
\midrule
 & The two readers & \multicolumn{2}{c}{64.2} & \\
\bottomrule
\end{tabular}
}
\end{table}

If the recording adds what the reader decides, speech models should predict the reading they hear rather than a reading of the text.
Some test utterances have the same text as an utterance read by another speaker in the development set, which no model has been trained on.
A text model receives the same input for both readings, so it can at best predict what the two readers have in common.
Table~\ref{tab:readers} compares each system's predictions with both readers.
The speech models agree with the reader they hear on \resval{rt-whisper-own}\% (Whisper) and \resval{rt-wavlm-own}\% (WavLM) of the words, but with the other reader only on \resval{rt-whisper-other}\% and \resval{rt-wavlm-other}\%.
This is close to how often the two readers agree with each other (\resval{rt-readers}\% on these pairs; \resval{agreement-3}\% over all texts read by two speakers).
The text classifiers and the prompted models show no such preference: they agree with the two readers about equally often, and slightly more with the other reader.
Although the sample is small, it does seem to support our interpretation of the gap between text and speech: the speech models do not only have better access to the same information, they follow the individual reading.

In fact, the text models match the reading of either speaker more often than the two speakers match each other (\resval{rt-modernbert-own}\% and \resval{rt-modernbert-other}\% for ModernBERT-large, against \resval{rt-readers}\%; Table~\ref{tab:readers}).
Therefore, the text model appears to predict something similar to the reading the speakers have in common, which is as close as text alone can get to any individual reading.

\section{Discussion}
\label{sec:discussion}

Why do prompted LLMs fall short of what their representations contain?
We see at least three possible reasons:
\begin{enumerate}\setlength{\itemsep}{1pt}\setlength{\parskip}{0pt}
\item The labels define prominence acoustically, as computed from one reader's voice, while the models' notion of emphasis may follow the meaning of the sentence.
\item A graded three-level scale is hard to express in words, while a classifier learns the cut-offs directly from the labels.
\item Numbering the words and producing a fixed answer format add errors of their own.
\end{enumerate}
The third reason is unlikely to explain much, since Qwen3.8 27B writes its answer under constrained decoding and still stays far below a linear probe on its representations.
The predicted label distributions speak against the second reason as well.
The best prompted model predicts the three classes in roughly the right proportions (\resval{profile-prompt-pred-0}, \resval{profile-prompt-pred-1} and \resval{profile-prompt-pred-2}\% against \resval{profile-prompt-gold-0}, \resval{profile-prompt-gold-1} and \resval{profile-prompt-gold-2}\% in the test set).
However, it finds only \resval{profile-prompt-recall-1}\% of the words labelled 1 and \resval{profile-prompt-recall-2}\% of those labelled 2, against \resval{profile-probe-recall-1}\% and \resval{profile-probe-recall-2}\% for the best linear probe.
The models therefore choose the wrong words rather than the wrong amount of emphasis.
Recalibrating the model's own label probabilities rules out the second reason more directly: the answer changes little whether it is recalibrated on 50 or 500 development utterances, and it stays far below the probe.
We believe that the first explanation is the most likely one: what the models report is a plausible reading of the sentence, built largely from what each word is usually like and partly from its context, not the reading in the recording.
Asking for the acoustic property instead, the words that would be read longest and loudest, does not help, so the model cannot simply be told which notion of prominence the labels use.
The explanation would also account for why examples help so slowly, since each example shows one reader's choices rather than a rule.
The size of the effect should be kept in proportion: the difference between the probe and the prompt (\resval{v2-e3-probe-prompt} points, 95\% interval \resval{v2-e3-probe-prompt-cilo}--\resval{v2-e3-probe-prompt-cihi}) is about a third of the difference that the reader makes to a speech model (\resval{v2-e3-reader-speech} points) and of the difference between speech and text (\resval{v2-e3-speech-text} points).

There is a parallel in human perception.
Listeners asked to mark prominent words rely not only on the acoustic signal but also on their expectations about the words, among them how frequent the words are \cite{cole2010,ladd2023}, and marks from untrained listeners become reliable only when those of about ten listeners are pooled \cite{cole2017}.
Explicit judgements of prominence thus draw on lexical expectations in humans too, and the prompted models' answers, which stay close to the word-majority lookup and use the context only in part, look more like such judgements than like the information a probe finds in their representations.
This is in line with the distinction between what models encode and what they report on metalinguistic questions \cite{hu2023}, and it suggests that asking an LLM about prosody tests its metalinguistic judgement rather than its knowledge of how sentences are read.

So, what is there in the text and what is there in the reading?
The ceiling we observed is empirical and applies to the models, data, and contexts we tested.
Preceding context, fine-tuning, and BiLSTM classifiers do not raise this limit, although this may partly reflect automatic labelling.
Given this, our results present a simple distinction.
The text largely determines prominence, in particular whether a word is prominent at all, and small encoders capture this as well as large LLMs do.
Where the preceding discourse calls for a contrast, LLMs can identify words to be emphasised from text \cite{turetzky2026}. However, in read audiobooks, much of the emphasis is not determined in this way.
The degree of emphasis received by content words and where readers place the main accent of a phrase vary from reader to reader and reading to reading, and this is what the recordings add: speech models follow individual readers, but text models do not (Table~\ref{tab:readers}).
However, since speech models are still \resval{toolkit-gap-3} points lower than the labelling procedure in the 3-way task, current speech representations do not even fully capture the acoustic part.

Under an independence assumption, the agreement between readers also bounds how high a text model could get at all.
If each reading of a text is a draw from a distribution $p$ over the labels a word can receive given the text, two readers give a word the same label with probability $\sum_l p_l^2$.
A text model, in contrast, can at best predict the most probable label, which is right with probability $\max_l p_l$.
Since $\sum_l p_l^2 \le \max_l p_l \le (\sum_l p_l^2)^{1/2}$, the agreement between readers bounds the best possible text-only accuracy from both sides: on average, it lies between \resval{bound-low-3}\% and \resval{bound-high-3}\% on the 3-way task and between \resval{bound-low-2}\% and \resval{bound-high-2}\% on the 2-way task (the upper bound by Jensen's inequality).
Resampling the \resval{bound-texts} texts changes the 3-way bounds little (95\% intervals \resval{bound-low-3-cilo}--\resval{bound-low-3-cihi}\% and \resval{bound-high-3-cilo}--\resval{bound-high-3-cihi}\%).
The best text models (\resval{text-best-3}\% and \resval{text-best-2}\%) lie within these bounds, in the lower half on the 3-way task.
The bounds assume that readers are independent draws given the text, although the realisation of intonation varies systematically with the context and between speakers \cite{arvaniti2024}, and since the labels are computed automatically, part of the disagreement between readers may come from the labelling rather than from the readers.
Neither assumption accounts for the distance between the text models and the upper bound.
Removing each speaker's average offset raises the 3-way bounds to \resval{v2-e4-spk-low}--\resval{v2-e4-spk-high}\%.
Correcting for label noise of the size the reconstruction suggests (\resval{v2-e4-noise-full-err}\% per label) raises the upper bound to \resval{v2-e4-noise-full-high}\%, but it raises the noise-free accuracy of the BiLSTM classifier on ModernBERT-large (on all test words) to \resval{v2-e4-noise-full-text}\% as well, so the distance between the two changes only from \resval{v2-e4-distance-observed} to \resval{v2-e4-distance-full} points.
The distance is larger for short utterances, on which text models are less accurate (\resval{v2-e4-short-text}\% at 5--10 words against \resval{v2-e4-long-text}\% at 21--40) while readers agree no less.

These results have practical consequences.
LLMs are increasingly asked to mark emphasis for speech synthesis or to pre-annotate speech data, while synthesised speech still conveys information status through intonation differently from human speakers \cite{hu2024tts}.
Asked directly, such a model will mostly reproduce a default placement of the main accent and miss many of the prominent words that its own representations encode, while a small classifier trained on labelled data recovers more, at a fraction of the cost.
Since text can only determine a reading, not the reading, systems that need to match a particular speaker will also need the speaker's own recordings.

Some limitations should be kept in mind.
Since the labels are produced automatically from the acoustics, speech models predict a function of their own input.
Part of the gap between text and speech may therefore reflect how the labels were made rather than how listeners perceive prominence; perceptual labels on a subset of the test set would show how large this effect is.
The labelling method itself was evaluated against manual accent annotation on the Boston University Radio News Corpus, where it was comparable with the best supervised methods \cite{suni2017}, but it has not been validated against listeners on LibriTTS, and listeners may judge as prominent the words that carry an accent rather than the most salient ones acoustically \cite{orrico2024metrical}.
Our word boundaries are not the ones the labels were computed on, but the reconstruction agrees with the original labels on \resval{toolkit-3}\% of the words, which suggests that the effect is small.
Our data are read audiobooks, all published before 1923 and ranging from verse with a fixed metre to children's stories, while the language models are trained on contemporary text.
Prominence may also be easier to predict from text in read speech than in conversation.
Our prompting results cover a limited set of instructions, answer formats and models, and the commercial models may change; asking for the acoustic property or for pairwise comparisons did not bring out more, but other elicitations might.
Finally, all our data are English, and languages in which prominence is more strongly tied to word order or morphology may leave less to the reader.

\section{Conclusion}

In this paper we have studied how much of prosodic prominence language models can find in text, whether they can report it, and what the speech signal adds.
Text models from BERT to LLMs with 30B parameters converge on the same accuracy, at most \resval{text-best-3}\% on the 3-way task, and classifiers trained on them share nearly half of their 2-way errors, which suggests that, for the models we tested, the ceiling lies in the task.
With the prompts we tested, LLMs do no better than a lexical baseline.
Their answers carry some information about context, but less than a linear probe finds in their contextual representations, and neither recalibration nor other ways of asking closes the gap.
Speech models add \resval{gap-3} points on the same words, mostly on content words and at the end of the utterance, and they are right on \resval{bd-3-all-wrong-audio}\% of the words that no text model gets right.
They also follow the individual reader: when two speakers read the same text, a speech model agrees with the one it hears far more often than with the other.
Overall, our results highlight the difference between what a language model encodes and what it reports.

Several questions remain open: whether the same results hold for perceptual labels and for conversational speech, whether speech-text LLMs can be fine-tuned to report what their encoders contain, and how much of the remaining variation speaker-aware models can predict.

\ifinterspeechfinal
\section{Acknowledgements}
The authors wish to acknowledge CSC -- IT Center for Science, Finland, for computational resources. The second author gratefully acknowledges support from the Human Diversity consortium, under the Profi7 program (grant no.~352727) and the Centre of Excellence (grant no.~374221) by the Research Council of Finland.
We thank Antti Suni for the Wavelet Prosody Toolkit and the prominence annotations of LibriTTS.
\fi

\bibliographystyle{IEEEtran}
\bibliography{refs}

\begin{thebibliography}{10}
\providecommand{\url}[1]{#1}
\csname url@samestyle\endcsname
\providecommand{\newblock}{\relax}
\providecommand{\bibinfo}[2]{#2}
\providecommand{\BIBentrySTDinterwordspacing}{\spaceskip=0pt\relax}
\providecommand{\BIBentryALTinterwordstretchfactor}{4}
\providecommand{\BIBentryALTinterwordspacing}{\spaceskip=\fontdimen2\font plus
\BIBentryALTinterwordstretchfactor\fontdimen3\font minus
  \fontdimen4\font\relax}
\providecommand{\BIBforeignlanguage}[2]{{%
\expandafter\ifx\csname l@#1\endcsname\relax
\typeout{** WARNING: IEEEtran.bst: No hyphenation pattern has been}%
\typeout{** loaded for the language `#1'. Using the pattern for}%
\typeout{** the default language instead.}%
\else
\language=\csname l@#1\endcsname
\fi
#2}}
\providecommand{\BIBdecl}{\relax}
\BIBdecl

\bibitem{talman2019}
A.~Talman, A.~Suni, H.~Celikkanat, S.~Kakouros, J.~Tiedemann, and M.~Vainio,
  ``Predicting prosodic prominence from text with pre-trained contextualized
  word representations,'' in \emph{Proceedings of the 22nd Nordic Conference on
  Computational Linguistics (NoDaLiDa)}, Turku, Finland, 2019, pp. 281--290.

\bibitem{ladd2023}
D.~R. Ladd and A.~Arvaniti, ``Prosodic prominence across languages,''
  \emph{Annual Review of Linguistics}, vol.~9, pp. 171--193, 2023.

\bibitem{hirschberg1993}
J.~Hirschberg, ``Pitch accent in context: Predicting intonational prominence
  from text,'' \emph{Artificial Intelligence}, vol.~63, no. 1--2, pp. 305--340,
  1993.

\bibitem{wagner2010}
M.~Wagner and D.~G. Watson, ``Experimental and theoretical advances in prosody:
  A review,'' \emph{Language and Cognitive Processes}, vol.~25, no. 7--9, pp.
  905--945, 2010.

\bibitem{bolinger1972}
D.~Bolinger, ``Accent is predictable (if you're a mind-reader),''
  \emph{Language}, vol.~48, no.~3, pp. 633--644, 1972.

\bibitem{yuan2005}
J.~Yuan, J.~M. Brenier, and D.~Jurafsky, ``Pitch accent prediction: Effects of
  genre and speaker,'' in \emph{Proc. Interspeech}, 2005, pp. 1409--1412.

\bibitem{li2025}
H.~Li, L.~Qu, J.~Hu, and T.~Li, ``{EME-TTS}: Unlocking the emphasis and emotion
  link in speech synthesis,'' in \emph{Proc. Interspeech}, 2025, pp.
  4368--4372.

\bibitem{rendel2016}
A.~Rendel, R.~Fernandez, R.~Hoory, and B.~Ramabhadran, ``Using continuous
  lexical embeddings to improve symbolic-prosody prediction in a text-to-speech
  front-end,'' in \emph{Proc. ICASSP}, 2016, pp. 5655--5659.

\bibitem{stephenson2022}
B.~Stephenson, L.~Besacier, L.~Girin, and T.~Hueber, ``{BERT}, can {HE} predict
  contrastive focus? predicting and controlling prominence in neural {TTS}
  using a language model,'' in \emph{Proc. Interspeech}, 2022, pp. 3383--3387.

\bibitem{gregory2004}
M.~L. Gregory and Y.~Altun, ``Using conditional random fields to predict pitch
  accents in conversational speech,'' in \emph{Proceedings of ACL}, 2004, pp.
  677--683.

\bibitem{nenkova2007}
A.~Nenkova, J.~Brenier, A.~Kothari, S.~Calhoun, L.~Whitton, D.~Beaver, and
  D.~Jurafsky, ``To memorize or to predict: Prominence labeling in
  conversational speech,'' in \emph{Proceedings of NAACL-HLT}, 2007, pp. 9--16.

\bibitem{stehwien2018}
S.~Stehwien, N.~T. Vu, and A.~Schweitzer, ``Effects of word embeddings on
  neural network-based pitch accent detection,'' in \emph{Proc. Speech
  Prosody}, 2018, pp. 719--723.

\bibitem{nielsen2020}
E.~Nielsen, M.~Steedman, and S.~Goldwater, ``The role of context in neural
  pitch accent detection in {E}nglish,'' in \emph{Proceedings of EMNLP}, 2020,
  pp. 7994--8000.

\bibitem{devlin2019}
J.~Devlin, M.-W. Chang, K.~Lee, and K.~Toutanova, ``{BERT}: Pre-training of
  deep bidirectional transformers for language understanding,'' in
  \emph{Proceedings of NAACL-HLT}, 2019, pp. 4171--4186.

\bibitem{kakouros2023bert}
S.~Kakouros and J.~O'Mahony, ``What does {BERT} learn about prosody?'' in
  \emph{Proceedings of the 20th International Congress of Phonetic Sciences
  (ICPhS)}, Prague, Czech Republic, 2023, pp. 1454--1458.

\bibitem{kakouros2023surprisal}
S.~Kakouros, J.~{\v{S}}imko, M.~Vainio, and A.~Suni, ``Investigating the
  utility of surprisal from large language models for speech synthesis
  prosody,'' in \emph{Proc. 12th ISCA Speech Synthesis Workshop (SSW)}, 2023,
  pp. 127--133.

\bibitem{suni2017}
A.~Suni, J.~{\v{S}}imko, D.~Aalto, and M.~Vainio, ``Hierarchical representation
  and estimation of prosody using continuous wavelet transform,''
  \emph{Computer Speech \& Language}, vol.~45, pp. 123--136, 2017.

\bibitem{kakouros2016}
S.~Kakouros and O.~R{\"a}s{\"a}nen, ``{3PRO} -- an unsupervised method for the
  automatic detection of sentence prominence in speech,'' \emph{Speech
  Communication}, vol.~82, pp. 67--84, 2016.

\bibitem{hsu2021}
W.-N. Hsu, B.~Bolte, Y.-H.~H. Tsai, K.~Lakhotia, R.~Salakhutdinov, and
  A.~Mohamed, ``{HuBERT}: Self-supervised speech representation learning by
  masked prediction of hidden units,'' \emph{IEEE/ACM Transactions on Audio,
  Speech, and Language Processing}, vol.~29, pp. 3451--3460, 2021.

\bibitem{chen2022}
S.~Chen, C.~Wang, Z.~Chen, Y.~Wu, S.~Liu, Z.~Chen, J.~Li, N.~Kanda,
  T.~Yoshioka, X.~Xiao, J.~Wu, L.~Zhou, S.~Ren, Y.~Qian, Y.~Qian, J.~Wu,
  M.~Zeng, X.~Yu, and F.~Wei, ``{WavLM}: Large-scale self-supervised
  pre-training for full stack speech processing,'' \emph{IEEE Journal of
  Selected Topics in Signal Processing}, vol.~16, no.~6, pp. 1505--1518, 2022.

\bibitem{radford2023}
A.~Radford, J.~W. Kim, T.~Xu, G.~Brockman, C.~McLeavey, and I.~Sutskever,
  ``Robust speech recognition via large-scale weak supervision,'' in
  \emph{Proceedings of ICML}, ser. Proceedings of Machine Learning Research,
  vol. 202, 2023, pp. 28\,492--28\,518.

\bibitem{delafuente2024}
A.~de~la Fuente and D.~Jurafsky, ``A layer-wise analysis of {M}andarin and
  {E}nglish suprasegmentals in {SSL} speech models,'' in \emph{Proc.
  Interspeech}, 2024, pp. 1290--1294.

\bibitem{wallbridge2025}
S.~Wallbridge, C.~Minixhofer, C.~Lai, and P.~Bell, ``Prosodic structure beyond
  lexical content: A study of self-supervised learning,'' in \emph{Proc.
  Interspeech}, 2025, pp. 4723--4727.

\bibitem{yosha2025whistress}
I.~Yosha, D.~Shteyman, and Y.~Adi, ``{WhiStress}: Enriching transcriptions with
  sentence stress detection,'' in \emph{Proc. Interspeech}, 2025, pp.
  4718--4722.

\bibitem{yosha2026stresstest}
I.~Yosha, G.~Maimon, and Y.~Adi, ``{StressTest}: Can {YOUR} speech {LM} handle
  the stress?'' in \emph{Findings of the Association for Computational
  Linguistics: ACL 2026}, 2026, pp. 1264--1285.

\bibitem{belinkov2022}
Y.~Belinkov, ``Probing classifiers: Promises, shortcomings, and advances,''
  \emph{Computational Linguistics}, vol.~48, no.~1, pp. 207--219, 2022.

\bibitem{hewitt2019}
J.~Hewitt and P.~Liang, ``Designing and interpreting probes with control
  tasks,'' in \emph{Proceedings of EMNLP-IJCNLP}, 2019, pp. 2733--2743.

\bibitem{brown2020}
T.~B. Brown, B.~Mann, N.~Ryder, M.~Subbiah, J.~Kaplan, P.~Dhariwal,
  A.~Neelakantan, P.~Shyam, G.~Sastry, A.~Askell, S.~Agarwal, A.~Herbert-Voss,
  G.~Krueger, T.~Henighan, R.~Child, A.~Ramesh, D.~M. Ziegler, J.~Wu,
  C.~Winter, C.~Hesse, M.~Chen, E.~Sigler, M.~Litwin, S.~Gray, B.~Chess,
  J.~Clark, C.~Berner, S.~McCandlish, A.~Radford, I.~Sutskever, and D.~Amodei,
  ``Language models are few-shot learners,'' in \emph{Advances in Neural
  Information Processing Systems}, vol.~33, 2020, pp. 1877--1901.

\bibitem{agarwal2024}
R.~Agarwal, A.~Singh, L.~Zhang, B.~Bohnet, L.~Rosias, S.~Chan, B.~Zhang,
  A.~Anand, Z.~Abbas, A.~Nova, J.~D. Co-Reyes, E.~Chu, F.~Behbahani, A.~Faust,
  and H.~Larochelle, ``Many-shot in-context learning,'' in \emph{Advances in
  Neural Information Processing Systems}, vol.~37, 2024, pp. 76\,930--76\,966.

\bibitem{hu2023}
J.~Hu and R.~Levy, ``Prompting is not a substitute for probability measurements
  in large language models,'' in \emph{Proceedings of EMNLP}, 2023, pp.
  5040--5060.

\bibitem{hu2024}
J.~Hu and M.~C. Frank, ``Auxiliary task demands mask the capabilities of
  smaller language models,'' in \emph{Proceedings of the Conference on Language
  Modeling (COLM)}, 2024, pp. 1--17.

\bibitem{wang2026}
Y.~S. Wang, L.~Qiu, Z.~Wu, R.~P. Levy, and Y.~Kim, ``Implicit representations
  of grammaticality in language models,'' in \emph{Proceedings of ACL (Volume
  1: Long Papers)}, 2026, pp. 15\,035--15\,055.

\bibitem{he2025}
L.~He, E.~Nie, H.~Schmid, H.~Schuetze, N.~Mesgarani, and J.~Brennan, ``Large
  language models as neurolinguistic subjects: Discrepancy between performance
  and competence,'' in \emph{Findings of the Association for Computational
  Linguistics: ACL 2025}, 2025, pp. 19\,284--19\,302.

\bibitem{azaria2023}
A.~Azaria and T.~Mitchell, ``The internal state of an {LLM} knows when it's
  lying,'' in \emph{Findings of EMNLP}, 2023, pp. 967--976.

\bibitem{burns2023}
C.~Burns, H.~Ye, D.~Klein, and J.~Steinhardt, ``Discovering latent knowledge in
  language models without supervision,'' in \emph{Proceedings of ICLR}, 2023,
  pp. 1--28.

\bibitem{zen2019}
H.~Zen, V.~Dang, R.~Clark, Y.~Zhang, R.~J. Weiss, Y.~Jia, Z.~Chen, and Y.~Wu,
  ``{LibriTTS}: A corpus derived from {LibriSpeech} for text-to-speech,'' in
  \emph{Proc. Interspeech}, 2019, pp. 1526--1530.

\bibitem{mcauliffe2017}
M.~McAuliffe, M.~Socolof, S.~Mihuc, M.~Wagner, and M.~Sonderegger, ``{Montreal
  Forced Aligner}: Trainable text-speech alignment using {Kaldi},'' in
  \emph{Proc. Interspeech}, 2017, pp. 498--502.

\bibitem{warner2024}
B.~Warner, A.~Chaffin, B.~Clavi{\'e}, O.~Weller, O.~Hallstr{\"o}m,
  S.~Taghadouini, A.~Gallagher, R.~Biswas, F.~Ladhak, T.~Aarsen, G.~T. Adams,
  J.~Howard, and I.~Poli, ``Smarter, better, faster, longer: A modern
  bidirectional encoder for fast, memory efficient, and long context finetuning
  and inference,'' in \emph{Proceedings of ACL (Volume 1: Long Papers)}, 2025,
  pp. 2526--2547.

\bibitem{conneau2020}
A.~Conneau, K.~Khandelwal, N.~Goyal, V.~Chaudhary, G.~Wenzek, F.~Guzm{\'a}n,
  E.~Grave, M.~Ott, L.~Zettlemoyer, and V.~Stoyanov, ``Unsupervised
  cross-lingual representation learning at scale,'' in \emph{Proceedings of
  ACL}, 2020, pp. 8440--8451.

\bibitem{cole2010}
J.~Cole, Y.~Mo, and M.~Hasegawa-Johnson, ``Signal-based and expectation-based
  factors in the perception of prosodic prominence,'' \emph{Laboratory
  Phonology}, vol.~1, no.~2, pp. 425--452, 2010.

\bibitem{cole2017}
J.~Cole, T.~Mahrt, and J.~Roy, ``Crowd-sourcing prosodic annotation,''
  \emph{Computer Speech \& Language}, vol.~45, pp. 300--325, 2017.

\bibitem{turetzky2026}
A.~Turetzky, A.~Dekel, H.~Aronowitz, R.~Hoory, and Y.~Adi, ``Knowing what to
  stress: A discourse-conditioned text-to-speech benchmark,'' in \emph{Proc.
  Interspeech}, 2026, pp. 2932--2936.

\bibitem{arvaniti2024}
A.~Arvaniti, A.~Katsika, and N.~Hu, ``Variability, overlap, and cue trading in
  intonation,'' \emph{Language}, vol. 100, no.~2, pp. 265--307, 2024.

\bibitem{hu2024tts}
N.~Hu, J.~Kim, R.~Orrico, S.~Gryllia, and A.~Arvaniti, ``Can {OpenAI}'s {TTS}
  model convey information status using intonation like humans?'' in
  \emph{Proc. Speech Prosody}, 2024, pp. 32--36.

\bibitem{orrico2024metrical}
R.~Orrico, S.~Gryllia, N.~Hu, and A.~Arvaniti, ``The role of metrical structure
  in prominence perception,'' in \emph{La voce nei media e nelle nuove
  tecnologie: produzione e percezione}, ser. Studi AISV, V.~{De Iacovo}, B.~M.
  {De Paolis}, and D.~Mereu, Eds.\hskip 1em plus 0.5em minus 0.4em\relax
  Milano: Officinaventuno, 2024, vol.~12, pp. 127--146.

\end{thebibliography}

\end{document}